\documentclass[letterpaper,english]{IEEEtran}
\usepackage{lmodern}

\usepackage[T1]{fontenc}
\usepackage[utf8]{inputenc}
\usepackage{color}
\usepackage{babel}
\usepackage{float}
\usepackage{textcomp}
\usepackage{mathrsfs}
\usepackage{url}
\usepackage{amsmath}
\usepackage{amsthm}
\usepackage{amssymb}
\usepackage{graphicx}
\usepackage[unicode=true]
 {hyperref}

\makeatletter

\pdfpageheight\paperheight
\pdfpagewidth\paperwidth

\floatstyle{ruled}
\newfloat{algorithm}{tbp}{loa}
\providecommand{\algorithmname}{Algorithm}
\floatname{algorithm}{\protect\algorithmname}

\theoremstyle{plain}
\newtheorem{thm}{\protect\theoremname}
\theoremstyle{definition}
\newtheorem{defn}[thm]{\protect\definitionname}
\theoremstyle{definition}
\newtheorem{example}[thm]{\protect\examplename}
\theoremstyle{plain}
\newtheorem{prop}[thm]{\protect\propositionname}
\theoremstyle{plain}
\newtheorem{cor}[thm]{\protect\corollaryname}

\usepackage{pdfsync}

\makeatother

\providecommand{\corollaryname}{Corollary}
\providecommand{\definitionname}{Definition}
\providecommand{\examplename}{Example}
\providecommand{\propositionname}{Proposition}
\providecommand{\theoremname}{Theorem}

\begin{document}
\title{Sum-product networks: A survey}
\author{Iago París, Raquel Sánchez-Cauce, Francisco Javier Díez\textit{\emph{$^{*}$}}\thanks{Department
of Artificial Intelligence, UNED, 28040 Madrid, Spain (e-mail: \{iagoparis,
rsanchez, fjdiez\}dia.uned.es). $^{*}$Corresponding author.\protect\\ Submitted to \textit{IEEE
Transactions on Pattern Analysis and Machine Intelligence}\textit{\emph{.
Comments are welcome.}}}}

\maketitle

\begin{abstract}
A sum-product network (SPN) is a probabilistic model, based on a rooted
acyclic directed graph, in which terminal nodes represent univariate
probability distributions and non-terminal nodes represent convex
combinations (weighted sums) and products of probability functions.
They are closely related to probabilistic graphical models, in particular
to Bayesian networks with multiple context-specific independencies.
Their main advantage is the possibility of building tractable models
from data, i.e., models that can perform several inference tasks in
time proportional to the number of links in the graph. They are somewhat
similar to neural networks and can address the same kinds of problems,
such as image processing and natural language understanding. This
paper offers a survey of SPNs, including their definition, the main
algorithms for inference and learning from data, the main applications,
a brief review of software libraries, and a comparison with related
models. 
\end{abstract}

\maketitle\thispagestyle{empty}

\section{Introduction}

\begin{figure}[h]
\begin{centering}
\includegraphics[scale=0.48]{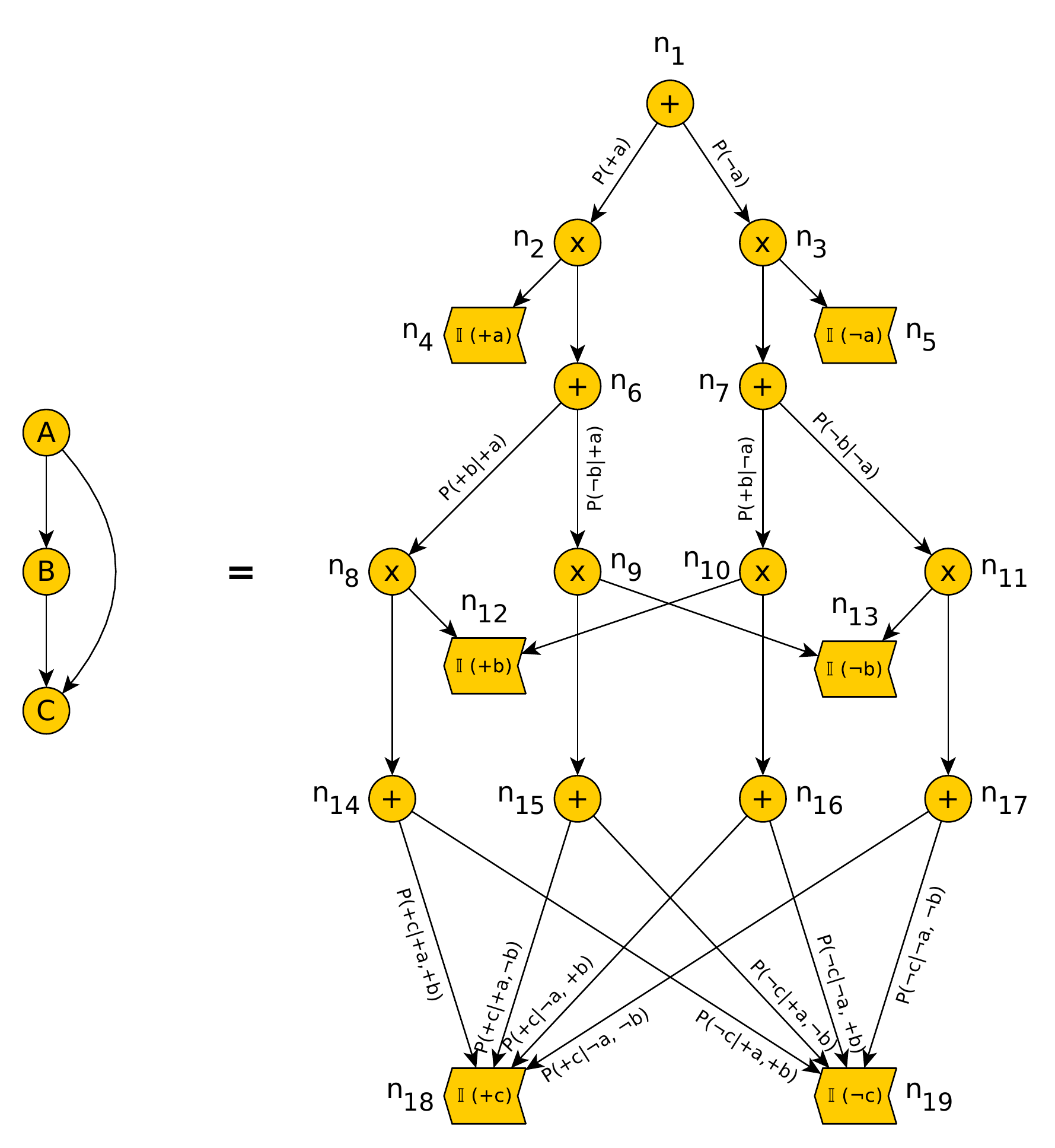}
\par\end{centering}
\caption{\label{fig:BN vs SPN}A Bayesian network (left) and an equivalent
SPN (right)\@. The~6 terminal nodes in the SPN are indicators for
the~3 variables in the model, $A$, $B$, and $C$; they are the
input of the SPN for every configuration of these variables, including
partial configurations. The root node, $n_{1}$, computes the joint
and marginal probabilities.}
\end{figure}

\IEEEPARstart{S}{um-product} networks (SPNs) were proposed by Poon
and Domingos \cite{poon2011} in 2011 as a modification of Darwiche's
\cite{darwiche2002,darwiche2003} arithmetic circuits. Every SPN consists
of a directed graph that represents a probability distribution resulting
from a hierarchy of distributions combined in the form of mixtures
(sum nodes) and factorizations (product nodes), as shown in Figure~\ref{fig:BN vs SPN}.
SPNs, like arithmetic circuits, can be built by transforming a probabilistic
graphical model \cite{koller2009}, such as a Bayesian network or
a Markov network, but they can also be learned from data. The main
advantage of SPNs is that several inference tasks can be performed
in time proportional to the number of links in the graph.

In this decade there has been great progress: numerous algorithms
have been proposed for inference and learning, and SPNs have been
successfully applied to many problems, including computer vision and
natural language processing, in which probabilistic models could not
compete with neural networks. The understanding of SPNs has also improved
and some aspect can now be explained more clearly than in the original
publications. For example, the first two papers about SPNs \cite{poon2011,gens2012}
presented them as an efficient representation of network polynomials,
while most of the later references, beginning with \cite{gens2013},
define them as the composition of probability distributions, which
is, in our view, more intuitive and much easier to understand. Consistency
was initially one of the defining properties of SPNs, which made them
more general than arithmetic circuits, but it later became clear that
decomposability, a stronger but much more intuitive property, suffices
to build SPNs for practical applications. In contrast, selectivity
(called determinism in arithmetic circuits), which was not mentioned
in the original paper \cite{poon2011}, proved to be relevant for
some inference tasks and for parameter learning \cite{peharz2017,peharz2015a}.
Additionally, some of the algorithms for SPNs are only sketched in
\cite{poon2011}, without much detail or formal proofs, and one of
them turned out to be correct only for selective SPNs. Other basic
algorithms are scattered over several papers, each using a different
mathematical notation.

For these reasons we decided to write a survey explaining the main
concepts and algorithms for SPNs with mathematical rigor. We have
intentionally avoided any reference to network polynomials, which
has forced us to develop new proofs for some algorithms and propositions,
alternative to those found in other references, such as \cite{peharz2015}.
We have also reviewed the literature on SPNs, with especial emphasis
on their applications. However, we are aware that some relevant references
may have been omitted, especially those published recently in this
thriving research area.

The rest of this paper is structured as follows. The two subsections
of this introduction highlight the significance of SPNs by comparing
them with probabilistic graphical models and neural networks respectively.
After some mathematical preliminaries (Sec.~\ref{sec:Preliminaries}),
we introduce the basic definitions of SPNs (Sec.\ref{sec:Definition})
and the main algorithms for inference (Sec.\ref{sec:Inference}),
parameter learning (Sec.~\ref{sec:Parameter-learning}), and structural
learning (Sec.~\ref{sec:Structure-learning}). We then review some
applications in several areas (Sec.~\ref{sec:Applications}), a few
open-source packages (Sec.~\ref{sec:Software}), and some extensions
of SPNs (Sec.~\ref{sec:Extensions}). Section~\ref{sec:Conclusion}
contains the conclusions, Appendix~A compares SPNs with arithmetic
circuits, Appendix~B analyzes the interpretation of sum nodes as
weighted averages of conditional probabilities, and Appendix~C contains
the proofs of all the propositions.

\subsection{SPNs vs.\ probabilistic graphical models (PGMs)\label{subsec:SPNs-vs-PGMs}}

\begin{figure}
\begin{centering}
\includegraphics[scale=0.52]{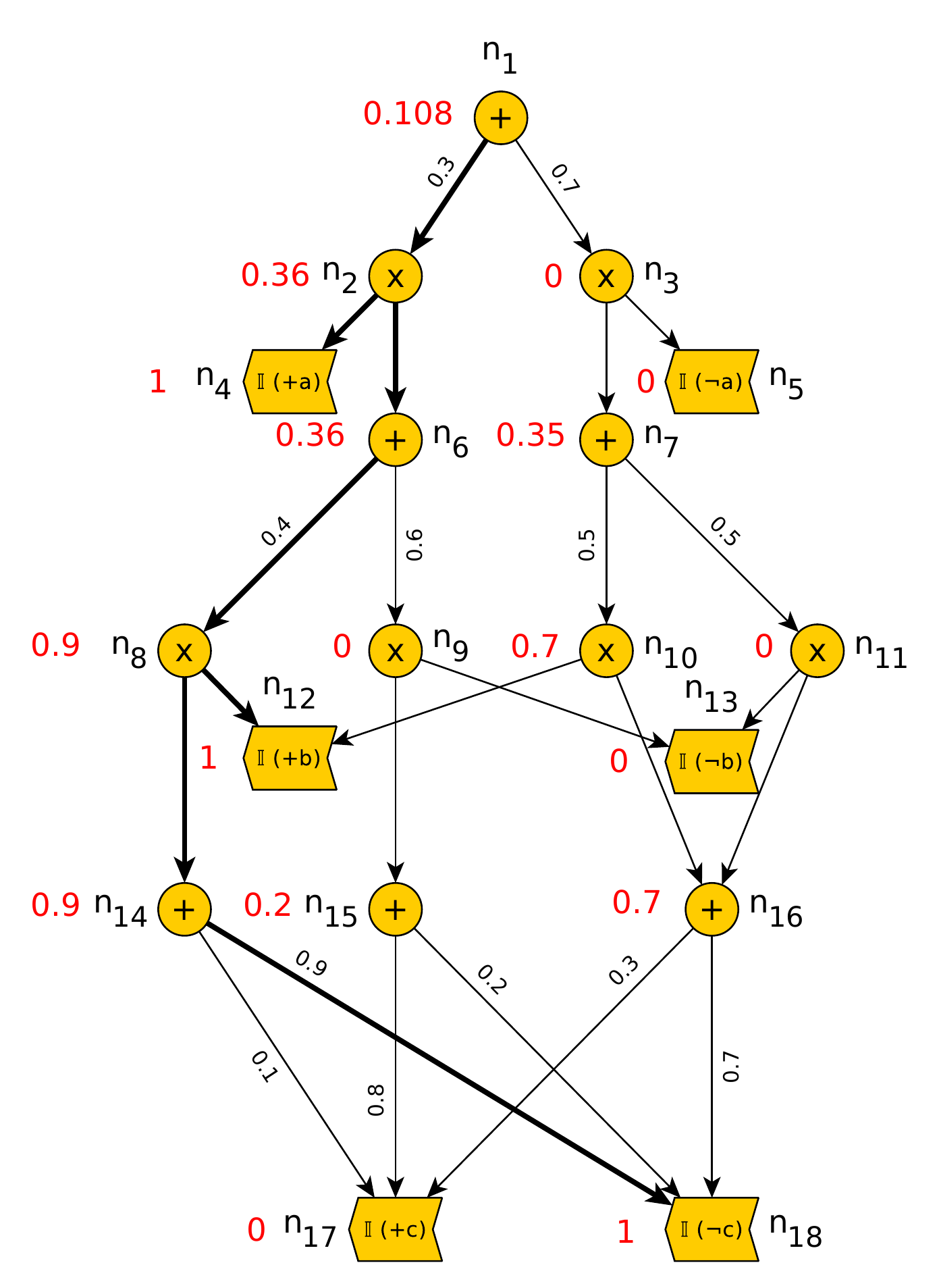}
\par\end{centering}
\caption{\label{fig:Inference SPN}If $P(c\,|\,b,+a)=P(c\,|\,b,\neg a)$ for
every value of~$B$ and~$C$ (context-specific independence), nodes~$n_{16}$
and~$n_{17}$ in Figure~\ref{fig:BN vs SPN} can be coalesced into
node~$n_{16}$ in this figure. The numbers in red are the values
$S_{i}(\mathbf{v})$ for $\mathbf{v}=(+a,+b,\neg c)$.}
\end{figure}

SPNs are similar to PGMs, such as Bayesian networks (BNs) and Markov
networks (also called Markov random fields) \cite{pearl1988,koller2009},
in their ability to compactly represent probability distributions.
The main difference is that in a PGM every node represents a variable
and---roughly speaking---links represent probabilistic dependencies,
sometimes due to causal influences, while in an SPN every node represents
a probability function. PGMs and other factored probability distributions
can be compiled into arithmetic circuits or SPNs \cite{chavira2007-spn-paper}.
In general a PGM is more compact than an equivalent SPN, as shown
in Figure~\ref{fig:BN vs SPN}. Inference in BNs and Markov networks
is an NP-problem and existing exact algorithms have worst-case exponential
complexity \cite{koller2009}, while SPNs can do inference in time
proportional to the number of links in the graph. In principle this
would not be an advantage for SPNs, because the conversion of a BN
or a Markov network into an SPN may create an exponential number of
links. However, context-specific independencies in BNs \cite{boutilier1996b-spn-paper}
can reduce the size of the corresponding SPN, as shown in Figure~\ref{fig:Inference SPN}.
In fact, any problem solvable in polynomial time with tabular BNs
(i.e., those whose conditional probabilities are given by tables)
can be solved in polynomial time with arithmetic circuits or SPNs
\cite{darwiche2003}, but the converse is not true---see the example
in \cite{poon2011}.\footnote{However, every SPN of finite-state variables can be converted into
a BN whose conditional probabilities are encoded as algebraic decision
diagrams (ADDs), in time and space proportional to the network size,
and it is possible to recover the original SPN with the variable elimination
algorithm \cite{zhao2015}\@.}

More importantly, while PGMs learned from data are usually intractable---except
for small problems or for specific types of models with limited expressiveness,
such as the naïve Bayes---the algorithms presented in Section~\ref{sec:Structure-learning}
can build tractable SPNs that yield excellent approximations both
for generative and discriminative tasks.

In contrast, BNs can be built using causal knowledge elicited from
human experts and there is a large body of recent research on building
causal BNs from experimental and/or observational data, under certain
conditions \cite{pearl2000}. It is also possible to combine causal
knowledge and data, and even to build BNs interactively \cite{bermejo2012a-spn-paper}.
All these options are currently impossible with SPNs. Additionally,
the independencies in a BN or in a Markov model are easier to read
than those in an SPN. On the other hand, the graph of an SPN can sometimes
be built from human knowledge to represent part/subpart and class/subclass
hierarchies---see \cite[Sec.~5]{poon2011} for an example.

In conclusion, each type of model has advantages and disadvantages,
and the choice for a real-world application must take into account
the size of the problem, the amount of knowledge and data available,
and the explanations required by the user.

\textcolor{magenta}{}

\subsection{SPNs vs.\ neural networks\label{subsec:SPNs-vs-NNs}}

SPNs can be seen as a particular type of feedforward neural networks
because there is a flow of information from the input nodes (the leaves)
to the output node (the root), but in this paper we reserve the term
``neural network'' (NN) for the models structured in layers connected
by the standard operators: sigmoid, ReLU, softmax, etc.

The main difference is that SPNs have a probabilistic interpretation
while standard NNs do not. Inference is also different: computing
a posterior probability requires two passes, and finding a the most
probable explanation (MPE) requires a backtrack from the root to the
leaves, as explained in Section~\ref{sec:Inference}. Additionally,
SPNs can do inference with partial information (i.e., when the values
of some of the variables are unknown), while in a NN it is necessary
to assign a value to each input node.

From the point of view of parameter learning, NNs are usually trained
with gradient descent or variations thereof, while SPNs can also be
trained with several probabilistic algorithms, such as EM and Bayesian
methods, which are much more efficient and have lower risk of overfitting
(cf.\ Sec.~\ref{sec:Parameter-learning}).

When building practical applications, the main difference is the possibility
of determining the structure of an SPN from data, looking for a balance
between model complexity and accuracy. In contrast, NNs are usually
designed by hand and it is necessary to examine different architectures
of different sizes with different hyperparameters, in a trial-and-error
approach, until a satisfactory model is found. For this reason, NNs
that have succeeded in practical applications are usually very big
and training them requires huge computational power. There are some
proposals to learn the structure of NNs using evolutionary computation,
which yields more efficient graphs, but this also requires immense
computational power \cite{stanley2002,ding2013}.

In spite of these advantages, NNs are still superior to SPNs in many
tasks. For example, in 2012 an SPN by Gens and Domingos \cite{poon2011}
achieved a classification accuracy of 84\% for the CIFAR-10 image
dataset, one of the highest scores at the time. However, deep NNs
have amply surpassed that result, reaching an impressive 99.3\% accuracy.\footnote{See \href{https://paperswithcode.com/sota/image-classification-on-cifar-10}{https://paperswithcode.com/sota/image-classification-on-cifar-10}.}

Nevertheless, in Section~\ref{sec:Applications} we mention several
applications in which SPNs are competitive with NNs and superior in
some aspects. For instance, random tensorized SPNs (RAT-SPNs) \cite{peharz2018}
have recently attained a classification accuracy comparable to deep
NNs for MNIST and other image datasets, with the advantages of being
interpretable as a probabilistic generative model and much more robust
to missing features. For another recent example, Stelzner et\ al.~\cite{stelzner2019}
proved that the attend-infer-repeat (AIR) framework used for object
detection and location is much more efficient when the variational
autoencoders (VAEs) are replaced by SPNs: they achieved an improvement
in speed of an order of magnitude, with slightly higher accuracy,
as well as robustness against noise. Other examples can be found in
Sections~\ref{sec:Applications} and~\ref{sec:Extensions}.

For a more detailed comparison of SPNs with NNs, VAEs, generative
adversarial networks (GANs), and other models, see \cite{peharz2018}.

\section{Mathematical preliminaries\label{sec:Preliminaries}}

In this paper we assume that every variable either has a finite set
of possible values, called \emph{states}, or is continuous, i.e.,
takes values in $\mathbb{R}$.

\subsection{Configurations of variables}

We denote by a capital letter, $V$, a variable and by the corresponding
lowercase letter, $v$, any value of $V$. Similarly a boldface capital
letter denotes a set of variables, $\mathbf{V}=\{V_{1},\ldots,V_{n}\}$,
the corresponding lowercase letter denotes any of its configurations,
$\mathbf{v}=(v_{1},\ldots,v_{n})$, and conf$(\mathbf{V})$ is the
set of all the configurations of $\mathbf{V}$. The empty set has
only one configuration, represented by $\blacklozenge$.

We denote by conf$^{*}(\mathbf{V})$ the set of all the configurations
of $\mathbf{V}$ and its subsets:
\begin{equation}
\text{conf}^{*}(\mathbf{V})=\{\mathbf{x}\mid\mathbf{X}\subseteq\mathbf{V}\}\;.
\end{equation}
Put another way,
\begin{equation}
\text{conf}^{*}(\mathbf{V})=\bigcup_{\mathbf{X}\in\mathscr{P}(\mathbf{V})}\text{conf}(\mathbf{X})
\end{equation}
We can think of $\text{conf}^{*}(\mathbf{V})\setminus\text{conf}(\mathbf{V})$
as the set of \emph{partial configurations} of~$\mathbf{V}$, i.e.,
the configurations in which only some of the variables in~$\mathbf{V}$
have an assigned value.

If $\mathbf{X}\subseteq\mathbf{V}$, the \emph{projection} (sometimes
called \emph{restriction}) of a configuration~$\mathbf{v}$ of~$\mathbf{V}$
onto~$\mathbf{X}$, $\mathbf{v}^{\downarrow\mathbf{X}}$, is the
configuration of $\mathbf{X}$ such that every variable $V\in\mathbf{X}$
takes the same value as in~$\mathbf{v}$; for example, $(+x_{1},+x_{2},\neg x_{3})^{\downarrow\{X_{1},X_{3}\}}=(+x_{1},\neg x_{3})$
and $(+x_{1},+x_{2},\text{\textlnot}x_{3})^{\downarrow\varnothing}=\blacklozenge$.
In order to simplify the notation, when~$\mathbf{X}$ has a single
variable, $V$, we will write $v$ instead of $(v)$ and $\mathbf{v}^{\downarrow V}$
instead of $\mathbf{v}^{\downarrow\{V\}}$; for example, $(+x_{1},+x_{2},\text{\textlnot}x_{3})^{\downarrow X_{2}}=+x_{2}$.

Given two configurations,~$\mathbf{x}$ and~$\mathbf{y}$, of two
disjoint sets,~$\mathbf{X}$ and~$\mathbf{Y}$, the \emph{composition}
of them, denoted by~$\mathbf{xy}$, is the configuration of $\mathbf{X}\cup\mathbf{Y}$
such that $(\mathbf{xy})^{\downarrow\mathbf{X}}=\mathbf{x}$ and $(\mathbf{xy})^{\downarrow\mathbf{Y}}=\mathbf{y}$.
For example, $(+x_{1},+x_{2})(\neg x_{3})=(+x_{1},+x_{2},\text{\textlnot}x_{3})$.

When $\mathbf{X}\subseteq\mathbf{V}$, a configuration~$\mathbf{x}$
is \emph{compatible} with configuration~$\mathbf{v}$ if $\mathbf{x}=\mathbf{v}^{\downarrow\mathbf{X}}$,
i.e., if every variable $V\in\mathbf{X}$ has the same value in both
configurations. All configurations are compatible with~$\blacklozenge$.
\begin{defn}
\label{def:indicator-FS}Given a value~$v$ of a finite-state variable~$V\in\mathbf{V}$,
we define the \emph{indicator function}, $\mathbf{\mathbb{I}}_{v}:\text{conf}(\mathbf{V})\mapsto\{0,1\}$,
as follows:
\begin{equation}
\mathbf{\mathbb{I}}_{v}(\mathbf{x})=\begin{cases}
1 & \text{if }V\notin\mathbf{X}\lor v=\mathbf{x}^{\downarrow V}\\
0 & \text{otherwise}\;.
\end{cases}\label{eq:indicator-FS}
\end{equation}
\end{defn}
If all the variables in~$\mathbf{V}$ are binary, then there are
$2n$ indicator functions.
\begin{example}
If $\mathbf{V}=\{V_{0},V_{1}\}$ and the domains of these variables
are $\{+v_{0},\neg v_{0}\}$ and $\{+v_{1},\neg v_{1}\}$ respectively,
then $\mathbf{\mathbb{I}}_{+v_{0}}(+v_{0},+v_{1})=1$, $\mathbf{\mathbb{I}}_{+v_{0}}(\neg v_{0},+v_{1})=0$,
$\mathbf{\mathbb{I}}_{+v_{0}}(+v_{1})=\mathbf{\mathbb{I}}_{+v_{0}}(\neg v_{1})=\mathbf{\mathbb{I}}_{+v_{0}}(\blacklozenge)=1$,
etc.
\end{example}

\subsection{Probability distributions and probability functions}
\begin{defn}
A \emph{probability distribution} defined on $\mathbf{V}$ is a function
$P:\text{conf}(\mathbf{V})\mapsto\mathbb{R}$ such that:
\begin{gather}
P(\mathbf{v})\geq0\;,\label{eq:P-non-negative}\\
\sum_{\mathbf{v}}P(\mathbf{v})=1\;.\label{eq:sum-1}
\end{gather}
\end{defn}
This definition can be extended so that $P$ represents not only a
probability distribution but also all its marginal probabilities,
as follows.
\begin{defn}
\label{def:prob-function}A \emph{probability function} defined on
$\mathbf{V}$ is a function $P:\text{conf}^{*}(\mathbf{V})\mapsto\mathbb{R}$
such that the restriction of~$P$ to $\text{conf}(\mathbf{V})$ is
a probability distribution and for every configuration~$\mathbf{x}$
such that $\mathbf{X}\subset\mathbf{V}$,
\begin{equation}
P(\mathbf{x})=\sum_{\mathbf{v}\mid\mathbf{v}^{\downarrow\mathbf{X}}=\mathbf{x}}P(\mathbf{v})\;.\label{eq:P-marginal}
\end{equation}
\end{defn}
This equation is the definition of marginal probability: $P(\mathbf{x})$
is obtained by summing the probabilities of all the configurations
of~$\mathbf{V}$ compatible with~$\mathbf{x}$.

\begin{prop}
\label{prop:marginal-distr}If $P$ is a probability function defined
on $\mathbf{V}$ and $\mathbf{X}\subset\mathbf{V}$, the restriction
of~$P$ to \textup{$\text{conf}(\mathbf{X})$} is a probability distribution
\textup{for~$\mathbf{X}$.}
\end{prop}
It is possible to define a new probability function as the sum or
the product of other probability functions.
\begin{prop}
\label{prop:sum-Pj}Let us consider $n$ probability functions $\{P_{1},\ldots,P_{n}\}$
defined on the same set of variables, $\mathbf{V}$, and $n$ weights,
$\{w_{1},\ldots,w_{n}\}$, with $w_{j}\geq0$ for every $j$, and
$\sum_{j=1}^{n}w_{j}=1$. The function $P:\text{conf}^{*}(\mathbf{V})\mapsto\mathbb{R}$,
such that for every configuration of $\mathbf{X}\subseteq\mathbf{V}$
\begin{equation}
P(\mathbf{x})=\sum_{j=1}^{n}w_{j}\cdot P_{j}(\mathbf{x})\;,
\end{equation}
 is a probability function. It is said to be a \emph{weighted average}
or a \emph{convex combination} of probability functions.
\end{prop}
\begin{prop}
\label{prop:prod-Pj}Let $\{P_{1},\ldots,P_{n}\}$ be a set of probability
functions defined on~$n$ disjoint sets of variables,~$\{\mathbf{V}_{1},\ldots,\mathbf{V}_{n}\}$,
respectively. Let $\mathbf{V}=\mathbf{V}_{1}\cup\ldots\cup\mathbf{V}_{n}$.
The function $P:\text{conf}^{*}(\mathbf{V})\mapsto\mathbb{R}$, such
that for every configuration of $\mathbf{X}\subseteq\mathbf{V}$
\begin{equation}
P(\mathbf{x})=\prod_{j=1}^{n}P_{j}(\mathbf{x})\;,
\end{equation}
is a probability function.
\end{prop}

\subsection{MAP, MPE, and MAX inference\label{subsec:MAP-MPE-MAX}}

In some inference tasks $\mathbf{e}$ denotes the evidence, i.e.,
the values observed (for example, the symptoms and signs of a medical
examination or the pixels in an image) and $\mathbf{X}$ the variables
of interest (for example, the possible diagnostics or the objects
that may be present in the image), with $\mathbf{X}\cap\mathbf{E}=\varnothing$.
In this context, $P(\mathbf{x}\mid\mathbf{e})$ is called the \emph{posterior
probability}.

The \emph{maximum a-posteriori} (MAP) configuration is 
\begin{equation}
\text{MAP}(\mathbf{e},\mathbf{X})=\underset{\mathbf{x}}{\arg\max}\:P(\mathbf{x}\,|\,\mathbf{e})\;.
\end{equation}
 Therefore, MAP inference divides the variables into three disjoint
sets: observed variables ($\mathbf{E}$), variables of interest ($\mathbf{X}$),
and hidden variables ($\mathbf{H}=\mathbf{V}\setminus(\mathbf{E}\cup\mathbf{X})$).

The \emph{most probable explanation} (MPE) is the configuration of
$\mathbf{X}=\mathbf{V}\setminus\mathbf{E}$ that maximizes the posterior
probability: 
\begin{equation}
\text{MPE}\,(\mathbf{e})=\underset{\mathbf{x}}{\arg\max}\:P(\mathbf{x}\,|\,\mathbf{e})\;.\label{eq:def-MPE}
\end{equation}
MPE is a special case of MAP in which $\mathbf{H}=\varnothing$, i.e.,
every variable that is not observed is a variable of interest. In
general, MAP inference is much harder than MPE \cite{park2002}.

Finally, MAX is a special case of MPE in which all the variables are
of interest, i.e., $\mathbf{X}=\mathbf{V}$ and $\mathbf{H}=\mathbf{E}=\varnothing$.
The MAX configuration is the configuration of $\mathbf{X}$ that maximizes
the probability: 
\begin{equation}
\text{MAX}\,(\mathbf{x})=\underset{\mathbf{x}}{\arg\max}\:P(\mathbf{x})\;.
\end{equation}

MPE and MAP are relevant when we wish to know the most probable configuration
of the variables of interest~$\mathbf{X}$ (for example, the possible
diagnostics), which is different from finding the most probable value
for each variable in~$\mathbf{X}$, as we will see in Example~\ref{example:MPE}.
MAP is relevant when some unobserved variables are not of interest;
for example,~$\mathbf{H}$ may represent the tests not performed:
these variables are neither observed nor part of the diagnosis. See
also \cite[Secs.~2.1.5.2 and~2.1.5.3]{koller2009}, where MPE and
MAP are called ``MAP'' and ``marginal MAP'' respectively. The
definition of MAX will be useful in Section~\ref{subsec:MAX-and-MAP}.

\subsection{Basic definitions about graphs}

Graphs have many applications in computer science. We describe here
the type of graph used to build SPNs.

A directed\emph{ graph} consists of a set of nodes and a set of directed
links. When there is a link $n_{i}\rightarrow n_{j}$ we say that~$n_{i}$
is a \emph{parent} of~$n_{j}$ and~$n_{j}$ is a \emph{child} of~$n_{i}$;
there cannot be another link from~$n_{i}$ to~$n_{j}$. Given a
node $n_{i}$, we denote by $pa(i)$ the set of indices of its parents
and by $ch(i)$ the set of indices of its children. For example, in
Figure~\ref{fig:BN vs SPN}, $ch(1)=\{2,3\}.$ Node~$n_{k}$ is
a \emph{descendant} of~$n_{i}$ if it is a child of~$n_{i}$ or
a child of a descendant of~$n_{i}$; we also say that $n_{i}$ is
an \emph{ancestor} of $n_{k}$. 

A cycle of length~$l$ consists of a set of~$l$ nodes and~$l$
links $\{n_{1}\rightarrow n_{2},n_{2}\rightarrow n_{3},\ldots,n_{l-1}\rightarrow n_{l},n_{l}\rightarrow n_{1}\}$.
A graph that contains no cycles, i.e., no node is a descendant of
itself, is \emph{acyclic}. An acyclic directed graph (ADG) is \emph{rooted}
if there is only one node (the \emph{root}, denoted by~$n_{r}$)
having no parents. \emph{Terminal nodes}, also called \emph{leaves},
are those that do not have children.

A \emph{directed tree} is a rooted ADG in which every node has one
parent, except the root. In this paper when we say ``a tree'' we
mean ``a directed tree''.

\section{Basic definitions of SPNs\label{sec:Definition}}

\subsection{Structure of an SPN}

An \emph{SPN} $\mathcal{S}$ consists of a rooted acyclic directed
graph such that:
\begin{itemize}
\item every leaf node represents a univariate probability distribution,
\item all the other nodes are either of type sum or product,
\item all the parents of a sum node (if any) are product nodes, and vice
versa, and
\item every link $n_{i}\rightarrow n_{j}$ outgoing from a sum node has
an associated weight, $w_{ij}\geq0$.
\end{itemize}
Usually $w_{ij}>0$. We will assume, unless otherwise stated, that
all SPNs are normalized, i.e., 
\[
\forall i,\;\sum_{j\in\text{\emph{ch}}(i)}w_{ij}=1\;.
\]

The probability distribution of each leaf node is defined on a variable~$V$,
which can have finite states or be continuous. In the first case,
the distribution is usually degenerate, i.e., there is a particular
value~$v^{*}$ of~$V$ such that $P(v^{*})=1$ and $P(v)=0$ otherwise.
In the graphical representation this leaf is denoted by the indicator
$\mathbb{I}_{v^{*}}$, sometimes written $\mathbb{I}(v^{*})$, as
in Figures~\ref{fig:BN vs SPN} and~\ref{fig:Inference SPN}. When~$V$
is continuous, the probability distribution can be Gaussian \cite{gens2013,rooshenas2014},
Poisson \cite{molina2017}, piecewise polynomial \cite{molina2018},
etc. (SPNs can be further generalized by allowing each terminal node
to represent a multivariate probability density---for example, a
multivariate Gaussian \cite{desana2017,hsu2017} or a Chow-Liu tree
\cite{vergari2015}.)

An SPN can be built bottom-up beginning with sub-SPNs of one node
and joining them with sum and product nodes. All the definitions of
SPNs can be established recursively, first for one-node SPNs, and
then for sum and product nodes. Similarly, all the properties of SPNs
can be proved by structural induction.

If the probability distribution of a leaf node is defined on~$V$,
its \emph{scope} is the set $\{V\}$. If~$n_{i}$ is a sum or a product
node, its \emph{scope} is the union of the scopes of its children:
\begin{equation}
\text{sc}(n_{i})=\bigcup_{j\in ch(i)}\text{sc}(n_{j})\;.
\end{equation}

The \emph{scope} of an SPN, denoted by $\text{sc}($$\mathcal{S}$),
is the scope of its root, $\text{sc}($$n_{r})$. The variables in
the scope of an SPN are sometimes called \emph{model variables}---in
contrast with \emph{latent variables}, which we present below. We
define conf($\mathcal{S}$) = conf($\text{sc}($$\mathcal{S}$)) and
conf$^{*}$($\mathcal{S}$) = conf$^{*}$($\text{sc}($$\mathcal{S}$)).

A sum node is \emph{complete} if all its children have the same scope.
An SPN is \emph{complete }if all its sum nodes are complete. (In arithmetic
circuits this property is called \emph{smoothness}.)

A product node is \emph{decomposable} if its children have disjoint
scopes. An SPN is \emph{decomposable} if all its product nodes are
decomposable.
\begin{prop}
\label{prop:decomp-node}A product node~$n_{i}$ is decomposable
if and only if no node in the SPN is a descendant of two different
children of~$n_{i}$.
\end{prop}
In the rest of the paper we assume that all the SPNs are complete
and decomposable.

\subsection{Node values and probability distributions\label{subsec:Node-values}}

In the following expressions we assume that all the variables have
finite states. If~$V$ were continuous, we would write $p(v)$ instead
of $P(v)$.
\begin{defn}[Value $S_{i}(\mathbf{x})$]
\label{def:value-Si} Let~$n_{i}$ be a node of~$\mathcal{S}$
and $\mathbf{x}\in\text{conf}^{*}(\mathcal{S})$. If~$n_{i}$ is
a leaf node that represents $P(v)$ then 
\begin{equation}
S_{i}(\mathbf{x})=\begin{cases}
P(\mathbf{x}^{\downarrow V}) & \text{if }V\in\mathbf{X}\\
1 & \text{otherwise}\;;
\end{cases}\label{eq:Si-leaf}
\end{equation}
if it is a sum node,
\begin{equation}
S_{i}(\mathbf{x})=\sum_{j\in ch(i)}w_{ij}\cdot S_{j}(\mathbf{x})\;,\label{eq:Si-sum}
\end{equation}
and if it is a product node, 
\begin{equation}
S_{i}(\mathbf{x})=\prod_{j\in ch(i)}S_{j}(\mathbf{x})\;.\label{eq:Si-prod}
\end{equation}
\end{defn}
Because of Equations~\ref{eq:indicator-FS} and~\ref{eq:Si-leaf},
if $P(v)$ is degenerate, with $P(v^{*})=1$, then $S_{i}(\mathbf{x})=\mathbb{I}_{v^{*}}(\mathbf{x})$,
which justifies using indicators as leaf nodes.
\begin{defn}[Value $S(\mathbf{x})$]
 The \emph{value} $S(\mathbf{x})$ returned by the SPN is the value
of the root, $S_{r}(\mathbf{x})$.
\end{defn}
\begin{thm}
\label{thm:Pi-Si}For a node $n_{i}$ in an SPN, the function $P_{i}:\text{conf}(\mathcal{S})\mapsto\mathbb{R}$,
such that 
\begin{equation}
P_{i}(\mathbf{x})=S_{i}(\mathbf{x})\;,\label{eq:Si=00003DPi}
\end{equation}
is a probability function defined on $\mathbf{V}=\text{sc}(n_{i})$.
\end{thm}
Please note that~$P_{i}$ is defined on the configurations of the
scope of the node, $\text{conf}^{*}(\text{sc}(n_{i}))$, while~$S_{i}$
is defined on the configurations of the scope of the whole network,
conf$^{*}$($\mathcal{S}$).

The probability function $P$ for the SPN is $P(\mathbf{x})=P_{r}(\mathbf{x})$.
The above theorem guarantees that the SPN properly computes a probability
distribution and all its marginal probabilities. The proof of the
theorem, in Appendix~C, relies on the completeness and decomposability
of the SPN\@. We offer there some counterexamples showing that if
an SPN is not complete or decomposable then $S(\mathbf{x})$ is not
necessarily a probability function. This theorem would still hold
if we replaced decomposability with a weaker condition, consistency
\cite{poon2011}, but this would complicate the definition of SPN
without offering any practical advantage.

\subsection{Selective SPNs}

We introduce now a particular type of SPNs that have interesting properties
for MPE inference and parameter learning and for the interpretation
of sum nodes.

When computing $S(\mathbf{x})$ for a given~$\mathbf{x}\in\text{conf}^{*}(\text{\ensuremath{\mathcal{S}}})$,
probability flows from the leaves to the root (cf.\ Def.~\ref{def:value-Si}).
Equation~\ref{eq:Si-sum} says that all the children of a sum node~$n_{i}$
can contribute to $S_{i}(\mathbf{x})$. However, $n_{i}$ may have
the property that for every configuration~$\mathbf{v}\in\text{conf}(\mathcal{S})$
at most one child makes a positive contribution, i.e., $S_{j}(\mathbf{x})=0$
for the other children of~$n_{i}$. We then say that~$n_{i}$ is
\emph{selective} \cite{peharz2014a}. The formal definition is as
follows.
\begin{defn}
\label{def:selective-node}A sum node~$n_{i}$ in an SPN is \emph{selective}
if
\begin{equation}
\forall\mathbf{v}\in\text{conf}(\mathcal{S}),\exists j^{*}\in\text{\emph{ch}}(i)\mid j\in\text{\emph{ch}}(i),j\neq j^{*}\Rightarrow S_{j}(\mathbf{v})=0\;.\label{eq:selective-node}
\end{equation}
\end{defn}
Please note that this definition says ``conf'', not ``conf$^{*}$''.
Therefore even if~$n_{i}$ is selective there may be a partial configuration
$\mathbf{x}\in\text{conf}^{*}(\mathcal{S})\setminus\text{conf}(\mathbf{V})$
such that several children of~$n_{i}$ make positive contributions
to $S_{i}(\mathbf{x})$.
\begin{defn}
\label{def:selective-SPN}An SPN is \emph{selective} if all its sum
nodes are selective.
\end{defn}
Arithmetic circuits satisfying this property are said to be \emph{deterministic}.
\begin{example}
Given the SPN in Figure~\ref{fig:Inference SPN}, we can check that
if $\mathbf{v}=(+a,+b,\neg c)$ then $S_{2}(\mathbf{v})=0.36$ and
$S_{3}(\mathbf{v})=0$. Only node~$n_{2}$ makes a positive contribution
to $S_{1}(\mathbf{v})$, so Property~\ref{eq:selective-node} holds
for this~$\mathbf{v}$ with $j^{*}=2$. We can make the same check
for each of the 6 sum nodes and each of the 8 configurations of $\{A,B,C\}$
in order to conclude that this SPN is selective. However, instead
of making these 48 checks, in the next section we show that for this
SPN selectivity is a consequence of the structure of its graph, regardless
of its numerical parameters (the weights).
\end{example}

\subsection{Sum nodes that represent model variables\label{subsec:Represent}}

Several papers about SPNs say that sum nodes represent latent random
variables. However, in this section we show that in some cases sum
nodes represent model variables.
\begin{defn}
\label{def:represents-V}Let~$n_{i}$ be a sum node having $m$ children
and~$V\in\text{sc}(n_{i})$ a variable with $m$ states. Let~$\sigma$
be a one-to-one function $\sigma:\{1,\ldots,m\}\mapsto ch(i)$. If
for every~$m\in\{1,\ldots,m\}$ either ~$\mathbb{I}_{v_{j}}$ is
a child of~$n_{\sigma(j)}$ (and hence a grandchild of~$n_{i}$)
or~$\mathbb{I}_{v_{j}}=n_{\sigma(j)}$ (i.e., the indicator itself
is a child of~$n_{i}$), we then say that ~$n_{i}$ \emph{represents}
variable~$V$.
\end{defn}
\begin{example}
Node~$n_{14}$ in Figure~\ref{fig:BN vs SPN} represents variable~$C$,
with $\sigma(1)=17$, $\sigma(2)=18$, because $\mathbb{I}_{c_{1}}=\mathbb{I}_{+c}=n_{17}=n_{\sigma(1)}$,
and $\mathbb{I}_{c_{2}}=\mathbb{I}_{\neg c}=n_{18}=n_{\sigma(2)}$.
Nodes~15, 16, and~17 also represent~$C$ for the same reason.

Node~$n_{6}$ represents variable~$B$ with $\sigma(1)=8$ and $\sigma(2)=9$
because ~$\mathbb{I}_{b_{1}}=\mathbb{I}_{+b}$ is a child of~$n_{\sigma(1)}=n_{8}$
and~$\mathbb{I}_{b_{2}}=\mathbb{I}_{\neg b}$ is a child of~$n_{\sigma(2)}=n_{9}$.
For analogous reasons node~$n_{7}$ also represents~$B$ and~$n_{1}$
represents~$A$.
\end{example}
In Appendix~B we prove that when the root node of an SPN represents
a variable~$V$, this node can be interpreted as the weighted average
of the conditional probabilities given~$V$, with $w_{i,\sigma(j)}=P(v_{j})$.
Similarly, if every ancestor of a sum node~$n_{i}$ represents a
variable, then~$n_{i}$ can be interpreted as the weighted average
of the conditional probabilities given the context defined by the
ancestors of~$n_{i}$.
\begin{prop}
\label{prop:represents-V}If a sum node~$n_{i}$ represents a model
variable~$V$, then $n_{i}$ is selective.
\end{prop}
This proposition is useful because it establishes a sufficient condition
for concluding that an SPN is selective by just observing its graph,
without analyzing its weights. For example, we can see that, according
to Definition~\ref{def:represents-V}, every node in Figures~\ref{fig:BN vs SPN}
and~\ref{fig:Inference SPN} represents a variable, which implies
that every node is selective and consequently both SPNs are selective.
In the next section we will use this property to transform any non-selective
SPN into selective.

\subsection{Augmented SPN\label{subsec:Augmented}}

\begin{figure}
\begin{centering}
\includegraphics[scale=0.45]{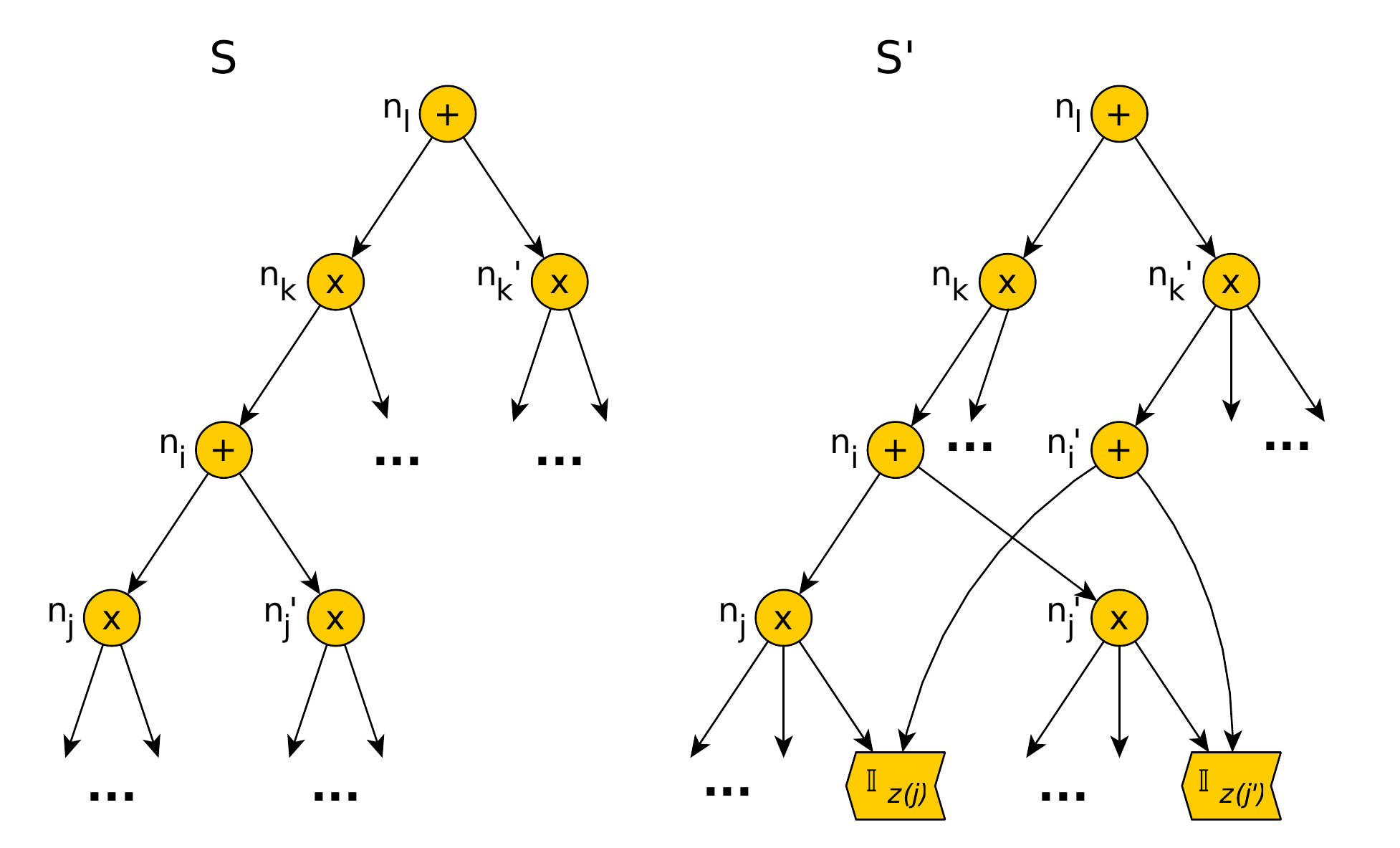}
\par\end{centering}
\caption{\label{fig:augmented-SPN}Augmentation of an SPN, assuming that $n_{i}$
is not selective in~$\mathcal{S}$\@. This process adds an indicator
$\mathbb{I}_{z(j)}$ for every child~$n_{j}$ of~$n_{i}$. Node~$n_{i'}$
is added to restore the completeness of~$n_{l}$ in~$\mathcal{S}'$.}
\end{figure}

The goal of augmenting a non-selective SPN~$\mathcal{S}$ \cite{peharz2015,peharz2017}
is to transform it into a selective SPN~$\mathcal{S}'$ that represents
the same probability function. For every non-selective node~$n_{i}$
in~$\mathcal{S}$ the augmentation consists in adding a new finite-state
variable~$Z$ so that $n_{i}$ represents~$Z$ in~$\mathcal{S}'$.
The process is as follows. For every child~$n_{j}$ we add a state
$z(j)$ to~$Z$. If~$n_{j}$ is a product node, we add the indicator~$\mathbb{I}_{z(j)}$
as a child of~$n_{j}$, as shown in Figure~\ref{fig:augmented-SPN};
if~$n_{j}$ is a terminal node, we insert a product node, make~$n_{j}$
a child of the new node (instead of being a child of $n_{i}$) and
add~$\mathbb{I}_{z(j)}$ as the second child of the new node. In
the resulting SPN,~$\mathcal{S}'$, $n_{i}$ represents variable~$Z$
(because of Def.~\ref{def:represents-V}) and is therefore selective.

However this transformation of the SPN may cause an undesirable side
effect. Let us assume, as shown in Figure~\ref{fig:augmented-SPN},
that~$n_{i}$ has a parent, $n_{k}$, and~$n_{l}$ is a parent of
both~$n_{k}$ and~$n_{k'}$. Even though~$n_{l}$ was complete
in~$\mathcal{S}$, the addition of~$Z$ has made this node incomplete
in~$\mathcal{S}'$ because $Z\in\text{sc}(n_{i})$ and $Z\in\text{sc}(n_{k})$
but $Z\notin\text{sc}(n_{k'})$. It is then necessary to make $Z\in\text{sc}(n_{k'})$
in order to restore the completeness of~$n_{l}$. So we create a
new sum node, $n_{i'}$, and make it a parent of all the indicators
of~$Z$, $\{\mathbb{I}_{z_{1}},\ldots,\mathbb{I}_{z_{m}}\}$ (see
again Fig.~\ref{fig:augmented-SPN}); the weights for~$n_{i'}$
can be chosen arbitrarily provided that they are all non-negative
and their sum is~1. If~$n_{k'}$ is a product node, then we add~$n_{i'}$
as a child of~$n_{k'}$. If~$n_{k'}$ is a terminal node, we insert
a product node, making both~$n_{i'}$ and~$n_{k'}$ children of
this new node. If~$n_{l}$ has other children, such as $n_{k''}$,
or ancestral sum nodes in~$\mathcal{S}$, then we must make each
one of them a parent or a grandparent of~$n_{i}'$, as we did for~$n_{k}'$.

Given that variable~$Z$ was not in the scope of the original SPN,
we can say that~$Z$ was \emph{latent} in~${\cal S}$ and the augmentation
of $n_{i}$ has made it explicit. The SPN~$\mathcal{S}'$ obtained
by augmenting all the non-selective nodes is said to be the \emph{augmented}
\emph{version} of~$\mathcal{S}$\@. Therefore, $\text{sc}(\mathcal{S}')=\text{sc}(\mathcal{S})\cup\mathbf{Z}$,
where~$\mathbf{Z}$ contains one variable for each sum node that
was not selective in~$\mathcal{S}$.\footnote{The original algorithm, proposed by Peharz's \cite{peharz2015}, augments
every node in~${\cal S}$, even those that were already selective.
In contrast, our algorithm only processes the nodes that were not
selective in~${\cal S}$, so that the augmentation of a selective
SPN does not modify it.}
\begin{prop}
\label{prop:augmented-SPN}If~$\mathcal{S}'$ is the augmented version
of~$\mathcal{S}$, then~$\mathcal{S}'$ is complete, decomposable,
and selective, and represents the same probability function for $\text{sc}(\mathcal{S})$,
i.e., if $\mathbf{x}\in\text{conf}^{*}(\mathcal{S})$, then $P'(\mathbf{x})=P(\mathbf{x})$.
\end{prop}

\subsection{Induced trees\label{subsec:induced-tree}}

This section is based on \cite{peharz2014a}.
\begin{defn}
\label{def:induced-SPN}Let $\mathcal{S}$ be an SPN and $\mathbf{v}\in\text{conf}(\mathcal{S})$
such that $S(\mathbf{v})\neq0$. The \emph{sub-SPN induced by~$\mathbf{v}$},
denoted by $\mathcal{S}_{\mathbf{v}}$, is a non-normalized SPN obtained
by (1) removing every node~$n_{i}$ such that $S_{i}(\mathbf{v})=0$
and the corresponding links, (2) removing every link $n_{i}\rightarrow n_{j}$
such that $w_{ij}=0$, and (3) removing recursively all the nodes
without parents, except the root.
\end{defn}
\begin{prop}
\label{prop:S_v(v)}If we denote by $S_{\mathbf{v}}(\mathbf{x})$
the value that~$\mathcal{S}_{\mathbf{v}}$ returns for~$\mathbf{x}$,
then $S_{\mathbf{v}}(\mathbf{v})=S(\mathbf{v})$.
\end{prop}
\begin{prop}
\label{prop:induced-tree}If $\mathcal{S}$ is selective, $\mathbf{v}\in\text{conf}(\mathbf{\mathcal{S}})$,
and $S(\mathbf{v})\neq0$, then $\mathcal{S}_{\mathbf{v}}$ is a tree
in which every sum node has exactly one child. Following the literature,
in this case we will write ${\cal T}_{\mathbf{v}}$ instead of $\mathcal{S}_{\mathbf{v}}$
to remark that it is a tree.
\end{prop}
\begin{example}
Given the SPN in Figure~\ref{fig:Inference SPN} and $\mathbf{v}=(+a,+b,\neg c)$,
$\mathcal{S}_{\mathbf{v}}$ only contains the links drawn with thick
lines in that figure and the nodes connected by them. This graph is
a tree because the SPN is selective.
\end{example}
When an SPN is selective, the set of trees obtained for all the configurations
in $\text{conf}(\mathbf{\mathcal{S}})$ is similar to the the set
of trees obtained by recursively decomposing the SPN, beginning from
the root, as proposed by Zhao et\ al.~\cite{zhao2016a}.
\begin{prop}
\label{prop:prod-w_ij} If $\mathcal{S}$ is selective, $\mathbf{v}\in\text{conf}(\mathbf{\mathcal{S}})$,
and $S(\mathbf{v})\neq0$ then 
\begin{equation}
S(\mathbf{v})=\prod_{(i,j)\in\mathcal{T_{\mathbf{v}}}}w_{ij}\prod_{n_{k}\text{ is terminal in }\mathcal{T_{\mathbf{v}}}}S_{k}(\mathbf{v})\;,\label{eq:prod-w_ij}
\end{equation}
 where $(i,j)$ denotes a link.
\end{prop}
\begin{cor}
\label{cor:prod-w_ij}If all the terminal nodes in~${\cal S}$ are
indicators, then
\[
S(\mathbf{v})=\prod_{(i,j)\in\mathcal{T_{\mathbf{v}}}}w_{ij}\;.
\]
\end{cor}
\begin{example}
For the SPN in Figure~\ref{fig:Inference SPN}, when $\mathbf{v}=(+a,+b,\neg c)$
we have $S(\mathbf{v})=w_{1,2}\cdot w_{6,8}\cdot w_{14,18}=0.3\cdot0.4\cdot0.9=0.108$.
\end{example}

\section{Inference \label{sec:Inference}}

\subsection{Marginal and posterior probabilities}

As defined in the previous section, $P(\mathbf{x})=S(\mathbf{x})=S_{r}(\mathbf{x})$.
The value $S(\mathbf{x})$ can be computed by an upward pass from
the leaves to the root in time proportional to the number of links
in the SPN. If $\mathbf{X}$ and $\mathbf{E}$ are two disjoint subsets
of~$\mathbf{V}$, then $P(\mathbf{x}\,|\,\mathbf{e})=S(\mathbf{xe})/S(\mathbf{e})$,
where $\mathbf{xe}$ is the composition of $\mathbf{x}$ and $\mathbf{e}$.
Therefore, any joint, marginal, or conditional probability can be
computed with at most two upward passes. Partial propagation, which
only propagates from the nodes in $\mathbf{X}\cup\mathbf{E}$, can
be significantly faster \cite{butz2018b}.

\subsection{MPE inference\label{subsec:MPE}}

The MPE configuration for an SPN is (see Sec.~\ref{subsec:MAP-MPE-MAX})
\begin{align}
\text{\emph{MPE}}\,(\mathbf{e}) & =\underset{\mathbf{x}}{\arg\max}\,P(\mathbf{x}\,|\,\mathbf{e})=\underset{\mathbf{x}}{\arg\max}\,P(\mathbf{xe})\nonumber \\
 & =\underset{\mathbf{x}}{\arg\max}\,S(\mathbf{xe})\;.\label{eq:MPE-S}
\end{align}

Let us assume that~$\mathcal{S}$ is selective. Then $\mathbf{X}\cup\mathbf{E}=\text{sc}(\mathcal{S})$
implies that $\mathbf{xe}\in\text{conf}(\mathcal{S}$) and, because
of Proposition~\ref{prop:induced-tree}, the sub-SPN induced by $\mathbf{xe}$
is a tree in which every sum node has only one child. Therefore, the
MPE can be found by examining all the trees for the configurations
$\mathbf{xe}$ in which~$\mathbf{e}$ is fixed and~$\mathbf{x}$
varies. It is possible to compare all those trees at once with a single
pass in~$\mathcal{S}$, by computing $S_{i}^{\text{max}}(\mathbf{e})$
for each node as follows:
\begin{itemize}
\item if $n_{i}$ is a sum node, then 
\begin{equation}
S_{i}^{\text{max}}(\mathbf{e})=\max_{j\in ch(i)}w_{ij}\cdot S_{j}^{\text{max}}(\mathbf{e})\;;
\end{equation}
\item otherwise $S_{i}^{\text{max}}(\mathbf{e})=S_{i}(\mathbf{e})$ (cf.\ Eqs.~\ref{eq:Si-leaf}
and~\ref{eq:Si-prod}).
\end{itemize}
Then the algorithm backtracks from the root to the leaves, selecting
for each sum node the child that led to $S^{\max}(\mathbf{e})$ and
for each product node all its children. When arriving at a terminal
node for variable~$V$, the algorithm selects the value $v=\underset{v'}{\arg\max}\,P(v')$.
In particular, if the terminal node is an indicator, $\mathbb{I}_{v^{*}}$,
then $v=v^{*}$. If~$V$ is continuous, then~$v$ is the mode of
$p(v)$. A tie means that there are two or more configurations having
the same probability $P(\mathbf{x},\mathbf{e})$; these ties can be
broken arbitrarily. The backtracking phase is equivalent to pruning
the SPN in order to obtain a tree in which every sum node has only
one child and there is exactly one terminal node for each variable;
the $v$ values selected at the terminal nodes make up the configuration
$\mathbf{\hat{x}}=\emph{MPE}\,(\mathbf{e})$.

\begin{figure}
\begin{centering}
\includegraphics[scale=0.52]{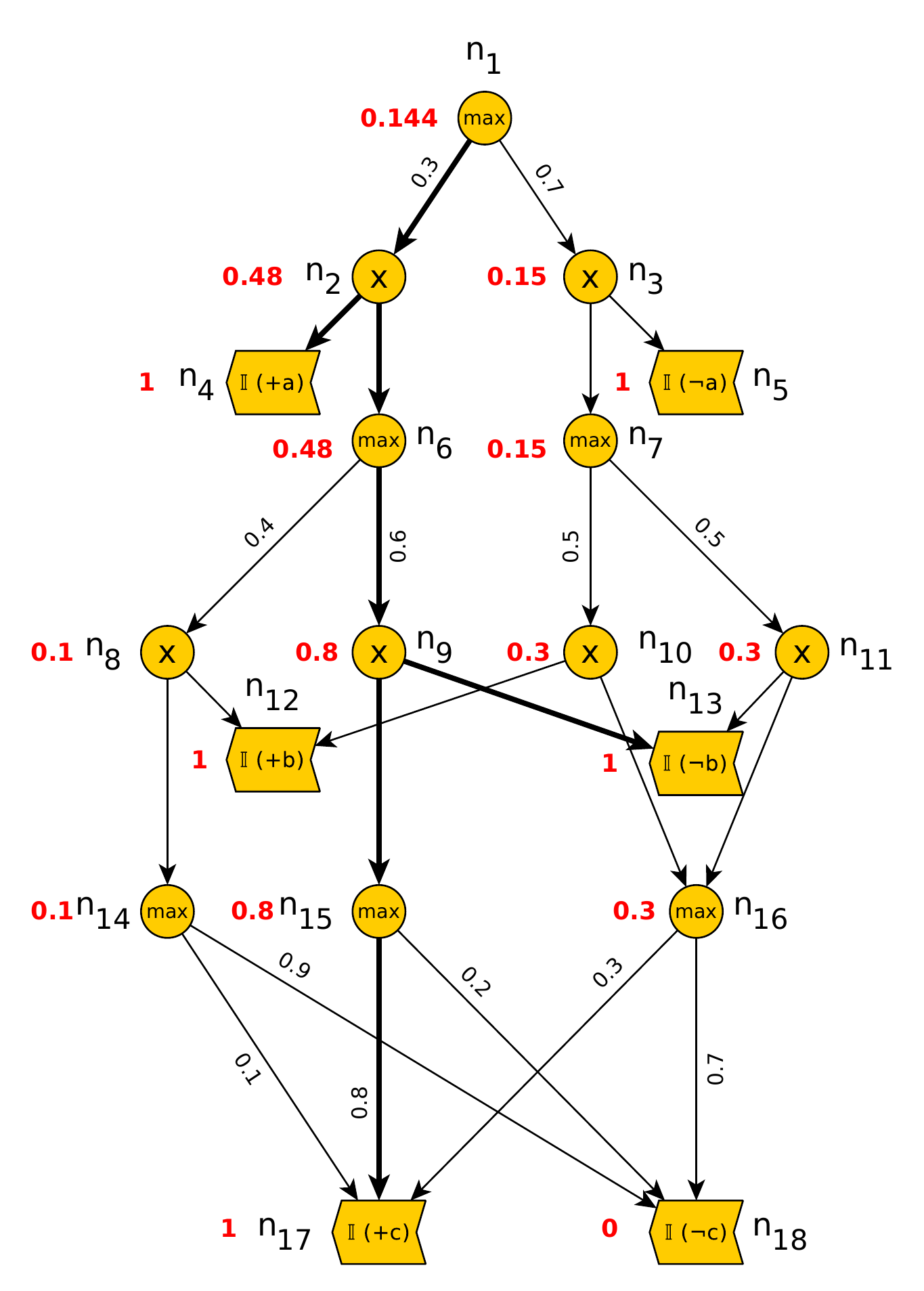}
\par\end{centering}
\caption{\label{fig:MPE SPN}MPE computation for the SPN in Figure~\ref{fig:Inference SPN}.
Sum nodes turn into max nodes. The numbers in red are the values $S_{i}^{\text{max}}(\mathbf{e})$
when the evidence is $\mathbf{\mathbf{e}}=+c$. The most probable
explanation, $\text{\emph{MPE}}\,(\mathbf{e})=(+a,\neg b)$, is found
by backtracking from the root to the leaves (thick lines).}
\end{figure}

\begin{example}
\label{example:MPE}Figure~\ref{fig:MPE SPN} shows the MPE inference
for the SPN in Figure~\ref{fig:Inference SPN} when $\mathbf{e}=+c$.
The MPE is obtained by backtracking from the root to the leaves: $\mathbf{\hat{x}}=\text{\emph{MPE}}\,(\mathbf{e})=(+a,\text{\textlnot}b)$.
We can check that $S_{1}^{\text{max}}(\mathbf{\mathbf{e}})=P(\mathbf{\hat{x}e})=P(+a,\neg b,+c)=0.144$.
For any other configuration $\mathbf{x}$ of $\mathbf{X}=\mathbf{V}\setminus\mathbf{E}=\{A,B\}$,
we have $S(\mathbf{xe})<S(\mathbf{\hat{x}e})$, in accordance with
Equation~\ref{eq:MPE-S}. We can also check that the nodes selected
by the backtracking phase are those of the tree induced by $\mathbf{\mathbf{\hat{x}}e}$---see
Def.~\ref{def:induced-SPN} and Prop.~\ref{prop:induced-tree}.

Note that, as mentioned in Section~\ref{subsec:MAP-MPE-MAX}, the
MPE cannot be determined by selecting the most probable value for
each variable. In this example $P(\neg a\,|\,\mathbf{e})=0.57>P(+a\,|\,\mathbf{e})$
and $P(\neg b\,|\,\mathbf{e})=0.68>P(+b\,|\,\mathbf{e})$, so we would
obtain the configuration $(\neg a,\neg b$), which is not the MPE.
\end{example}
This algorithm was proposed by Chan and Darwiche \cite{chan2006}
for arithmetic circuits, adapted to SPNs by Poon and Domingos \cite{poon2011},
and later called Best Tree (BT) in \cite{mei2018}. Peharz \cite[Theorem~2]{peharz2017}
proved that when an SPN is selective, BT computes the true MPE\@.
However, when a network is not selective, the sub-SPN induced by a
configuration $\mathbf{xe}$ is not necessarily a tree, so the value
$S^{\max}(\mathbf{e})$ computed by BT---which only considers the
probability that flows along trees with one child for each sum node---may
be different from $\max_{\mathbf{x}}S(\mathbf{xe})$ and, consequently,
the configuration returned by BT may be different from the true MPE\@.
Therefore, even though the MPE can be found in time proportional to
the size of the graph for selective SPNs, MPE is NP-complete for general
SPNs \cite[Theorem~5.3]{peharz2015}.\textcolor{magenta}{}\footnote{When an SPN $\mathcal{S}$ is not selective, it is possible find an
approximation to the MPE by augmenting it and then finding the MPE
for $\mathcal{S}'$ given~$\mathbf{e}$. The result is a configuration~$\mathbf{y}$
of $\mathbf{\mathbf{Y}}=(\mathbf{V}\cup\mathbf{Z})\setminus\mathbf{E}=\mathbf{X}\cup\mathbf{Z}$,
which we can then project onto~$\mathbf{X}$. However, Park \cite{park2002}
proved that in general this method does not find good approximations,
i.e., the posterior probability of the configuration found by this
method may differ significantly from that of the true MPE\@.}

\subsection{MAX and MAP\label{subsec:MAX-and-MAP}}

Exact MAP inference for SPNs is NP-hard because it includes as a particular
case MPE (see Sec.~\ref{subsec:MAP-MPE-MAX}), which is NP-complete.
Nevertheless, Mei et\ al.~\cite{mei2018} proposed several algorithms
that are very efficient in practice. First, they presented an algorithm
for the MAX problem in general SPNs. Second, they proved that every
MAP problem for SPNs can be reduced to a MAX problem for a new SPN
built in linear time. This way they were able to exactly solve MAP
problems for SPNs with up to $1,000$ variables and $150,000$ links.

Third, they proposed several approximate MAP solvers that trade accuracy
for speed, obtaining excellent results. In particular, they extended
the BT method to the MAX problem for non-selective SPNs. This extension,
called $K$-Best Tree (KBT), selects the~$K$ trees with the largest
output. Then, the corresponding configurations are obtained (by backtracking)
and evaluated in the SPN. The one with the largest output is the approximate
solution to the MAX problem. Note that, for $K=1$, KBT reduces to
BT.

\section{Parameter learning \label{sec:Parameter-learning}}

Parameter learning consists in finding the optimal parameters for
an SPN given its graph and a dataset. In \emph{generativ}e learning
the most common optimality criterion is to maximize the likelihood
of the parameters of the model given a dataset, while in \emph{discriminative}
learning the goal is to maximize the conditional likelihood for each
value of a variable~$C$, called the \emph{class}.

\subsection{Maximum likelihood estimation (MLE)\label{subsec:MLE}}

Let $\mathcal{D}=\{\mathbf{v}^{1},\mathbf{v}^{2},\ldots,\mathbf{v}^{T}\}$
be a dataset of~$T$ independent and identically distributed (i.i.d.)
instances. We denote by $\mathbf{W}$ the set of weights of the SPN
and by $\mathbf{\Theta}$ the parameters of the probability distributions
in the terminal nodes; both of them act as conditioning variables
for the probability of the instances in the database. We define~$L_{\mathcal{D}}(\mathbf{w},\boldsymbol{\theta})$
as the logarithm of the likelihood: 
\begin{align}
L_{\mathcal{D}}(\mathbf{w},\boldsymbol{\theta}) & =\log P(\mathcal{D}\,|\,\mathbf{w},\boldsymbol{\theta})\nonumber \\
 & =\sum_{t=1}^{T}\log S(\mathbf{v}^{t}\,|\,\mathbf{w},\boldsymbol{\theta})\;.\label{eq:L_D}
\end{align}
The goal is to find the configuration of $\mathbf{W}\cup\mathbf{\Theta}$
that maximizes $L_{\mathcal{D}}(\mathbf{w},\boldsymbol{\theta})$.
Given that there is no restriction linking the parameters of one node
(either sum, product, or terminal) with those of others, the optimization
can be done independently for each node. For terminal nodes representing
univariate distributions, standard statistical techniques can be applied.
In this section we will focus on the optimization of the weights,~$\mathbf{W}$,
so we omit~$\mathbf{\Theta}$ in the equations. The configuration
that maximizes the likelihood is
\begin{align}
\widehat{\mathbf{w}} & =\underset{\mathbf{w}}{\arg\max}\,P(\mathcal{D}\,|\,\mathbf{w})=\underset{\mathbf{w}}{\arg\max}\,L_{\mathcal{D}}(\mathbf{w})\label{eq:argmax-likelihood}
\end{align}
 subject to $w_{ij}\geq0$ and $\sum_{j\in ch(i)}w_{ij}=1$.

\subsubsection{MLE for selective SPNs\label{subsec:max-likelihood-selective}}

When the SPN is selective and $S(\mathbf{v})\neq0$, then the weights
of the sum nodes can be estimated in closed form by applying MLE \cite{peharz2014a}
as follows.
\begin{prop}
\label{prop:nij-log-wij}When an SPN is selective and its weights
are all different from~0,
\begin{equation}
L_{\mathcal{D}}(\mathbf{w})=\sum_{i}\sum_{j\in\text{ch}(i)}n_{ij}\cdot\log w_{ij}+c\;,\label{eq:nij-log-wij}
\end{equation}
where $n_{ij}$ is the number of instances in the dataset for which
$(i,j)\in\mathcal{T}_{\mathbf{v}^{t}}$ and~$c$ is a value that
does not depend on~$\mathbf{w}$.
\end{prop}
The $n_{ij}$'s can be computed by having a counter for every link
in the SPN\@. For each instance $\mathbf{v}^{t}$ in the dataset,
we compute $S(\mathbf{v}^{t})$ and then backtrack from the root to
the leaves: for each product node we select all its children and for
each sum node~$n_{i}$ we select the only child for which $S_{j}(\mathbf{v}^{t})>0$
and increase by~1 the counter $n_{ij}$.

It is then necessary to obtain the configuration~$\widehat{\mathbf{w}}$
that maximizes the likelihood, defined in Equation~\ref{eq:argmax-likelihood}.
The only constraint is $\sum_{j\in ch(i)}w_{ij}=1$ for every~$i$,
which implies that the parameters for one node can be optimized independently
of those for other nodes. The values that maximize the $i$-th term
in Equation~\ref{eq:nij-log-wij} are
\begin{equation}
\widehat{w}_{ij}=\dfrac{n_{ij}}{\sum_{j'\in ch(i)}n_{ij'}}\;.\label{eq:estimation-w_ij}
\end{equation}
There is a special case in which $\sum_{j\in ch(i)}n_{ij}=0$. This
occurs when $S_{i}(\mathbf{v}^{t})=0$ for every~$t$, i.e., when
none of the instances in the dataset propagates through the sum node
$n_{i}$. In this case, the weights of this node can be set uniformly:
\[
\forall j\in ch(i),\;\widehat{w}_{ij}=\dfrac{1}{|ch(i)|}\;.
\]
Alternatively, it is possible to use a Laplace-like smoothing parameter~$\alpha$,
so that
\begin{equation}
\widehat{w}_{ij}=\dfrac{n_{ij}+\alpha}{\sum_{j'\in ch(i)}(n_{ij'}+\alpha)}\;,\label{eq:estimation-w_ij-Laplace}
\end{equation}
with $0<\alpha\leq1$.

\subsubsection{Partial derivatives of~$S$}

For every node~$n_{i}$ we define
\begin{equation}
S_{i}^{\partial}(\mathbf{x})=\frac{1}{S(\mathbf{x})}\cdot\frac{\partial S}{\partial S_{i}}(\mathbf{x})\;.\label{eq:S-deriv-i}
\end{equation}
For the root node we have 
\begin{equation}
S_{r}^{\partial}(\mathbf{x})=\frac{1}{S(\mathbf{x})}\cdot\frac{\partial S}{\partial S_{r}}(\mathbf{x})=\frac{1}{S(\mathbf{x})}\;.\label{eq:S-deriv-r}
\end{equation}
If~$n_{j}$ is not the root,
\begin{align*}
S_{j}^{\partial}(\mathbf{x}) & =\frac{1}{S(\mathbf{x})}\cdot\frac{\partial S}{\partial S_{j}}(\mathbf{x})\\
 & =\frac{1}{S(\mathbf{x})}\cdot\sum_{i\in pa(j)}\frac{\partial S}{\partial S_{i}}(\mathbf{x})\cdot\frac{\partial S_{i}}{\partial S_{j}}(\mathbf{x})\\
 & =\sum_{i\in pa(j)}S_{i}^{\partial}(\mathbf{x})\cdot\frac{\partial S_{i}}{\partial S_{j}}(\mathbf{x})\;,
\end{align*}
where $pa(j)$ is the set of indices for the parents of~$n_{j}$.
If~$n_{i}$ is a sum node, 
\begin{equation}
S_{j}^{\partial}(\mathbf{x})=\sum_{i\in pa(j)}w_{ij}\cdot S_{i}^{\partial}(\mathbf{x})\;;\label{eq:deriv-sum}
\end{equation}
if it is a product node, 
\begin{equation}
S_{j}^{\partial}(\mathbf{x})=\sum_{i\in pa(j)}S_{i}^{\partial}(\mathbf{x})\cdot\prod_{\substack{j'\in ch(i)\setminus\{j\}}
}S_{j'}(\mathbf{x})\;.\label{eq:deriv-prod}
\end{equation}
These equations mean that, for every node~$n_{j}$, $S_{j}^{\partial}(\mathbf{x})$
can be computed once we have the $S_{i}^{\partial}(\mathbf{x})$'s
of its parents and the $S_{j'}(\mathbf{x})$'s of its siblings. Therefore,
after computing $S_{i}(\mathbf{x})$ for every node with an upward
pass, $S_{i}^{\partial}(\mathbf{x})$ can be computed with a downward
pass, both in linear time. This algorithm is similar to backpropagation
for neural networks and can be implemented using software packages
that support automatic differentiation.

\subsubsection{Gradient descent}

\paragraph{Standard gradient descent}

Gradient descent (GD), a well known optimization method, was proposed
for SPNs for both generative and discriminative models in \cite{poon2011}
and \cite{gens2012}, respectively.\footnote{The method is commonly called ``gradient descent'' when its goal
is to minimize a quantity---for example, the classification error
in neural networks. In our case it would be more appropriate to call
it ``gradient \emph{ascent}'' because the goal is to maximize the
likelihood. However, in this paper we follow the standard terminology
for SPNs.} The algorithm is initialized by assigning an arbitrary value to each
parameter, $\widehat{w}_{ij}^{(0)}$. and in every iteration,~$s$,
this value is updated, in order to increase the likelihood of the
model: 
\begin{equation}
\widehat{w}_{ij}^{(s+1)}=\widehat{w}_{ij}^{(s)}+\gamma\frac{\partial L_{\mathcal{D}}(\mathbf{w})}{\partial w_{ij}}\;,
\end{equation}
where $\gamma$ is the learning rate (a hyperparameter). It may be
necessary to renormalize the weights for each~$i$ after each update.

The partial derivative for $L_{{\cal D}}$ can be efficiently computed
using the $S_{i}^{\partial}$'s defined above, as follows. Because
of Equation~\ref{eq:L_D},

\[
\frac{\partial L_{\mathcal{D}}(\mathbf{w})}{\partial w_{ij}}=\sum_{t=1}^{T}\frac{\partial\log S}{\partial w_{ij}}(\mathbf{v}^{t})\;,
\]
and
\begin{align*}
\frac{\partial\log S}{\partial w_{ij}}(\mathbf{v}^{t}) & =\frac{1}{S(\mathbf{v}^{t})}\cdot\frac{\partial S}{\partial w_{ij}}(\mathbf{v}^{t})\\
 & =\frac{1}{S(\mathbf{v}^{t})}\cdot\frac{\partial S}{\partial S_{i}}(\mathbf{v}^{t})\cdot\frac{\partial S_{i}}{\partial w_{ij}}(\mathbf{v}^{t})\;,
\end{align*}
which together with Equations~\ref{eq:Si-sum} and~\ref{eq:S-deriv-i}
leads to 
\[
\frac{\partial\log S}{\partial w_{ij}}(\mathbf{v}^{t})=S_{i}^{\partial}(\mathbf{v}^{t})\cdot S_{j}(\mathbf{v}^{t})
\]
and, finally,
\begin{equation}
\frac{\partial L_{\mathcal{D}}(\mathbf{w})}{\partial w_{ij}}=\sum_{t=1}^{T}S_{i}^{\partial}(\mathbf{v}^{t})\cdot S_{j}(\mathbf{v}^{t})\;.
\end{equation}
This equation allows us to perform each iteration of GD in time proportional
to the size of the SPN and the number of instances in the database.

\paragraph{Stochastic GD and mini-batch GD}

In the stochastic version of GD, each iteration~$s$ uses only one
instance of the database, randomly chosen, so that
\[
\widehat{w}_{ij}^{(s+1)}=w_{ij}^{(s)}+\gamma\dfrac{\partial\log S}{\partial w_{ij}}(\mathbf{v}^{t})
\]
until the algorithm converges. 

Another possibility is to use mini-batches to update the weights,
that is, to divide the dataset in batches of~$L$ randomly drawn
instances, where $L<T$ (usually $L\ll T$), and update the weights
as follows:
\[
\widehat{w}_{ij}^{(s+1)}=\widehat{w}_{ij}^{(s)}+\gamma\sum_{l=1}^{L}\frac{\partial\log S}{\partial w_{ij}}(\mathbf{v}^{l})\;.
\]
This version is the most popular when applying GD\@.

\paragraph{Hard GD}

The application of GD to deep networks, either neural networks or
SPNs, suffers from the vanishing gradients problem: the deeper the
layer, the lower the contribution of its weights to the model output,
so the influence of the parameters in the deepest layers may be imperceptible.
The \emph{hard} version of GD for SPNs solves this problem by replacing
the sum nodes with max nodes and reparametrizing the weights so that
the gradient of the log-likelihood function remains constant. This
method was introduced for SPNs by Gens and Domingos \cite{gens2012}
for discriminative learning.

\subsubsection{Expectation-Maximization (EM)\label{subsec:EM}}

\paragraph{Standard EM}

We have seen how to learn the parameters of a selective SPN from a
complete dataset. However, many real-world problems have missing values.
We denote by~$\mathbf{H}^{t}$ the variables missing (hidden) in
the $t$-th instance of the database. Additionally, when learning
the parameters of~${\cal S}'$, an augmented SPN, the database is
always incomplete, even if it contains all the values for the the
model variables in ${\cal S}$, because it does not contain the latent
variables~$\mathbf{Z}$, added when augmenting the SPN, so $\mathbf{Z}\subseteq\mathbf{H}^{t}$
for every~$t$.

In this situation we can apply the expectation-maximization (EM) algorithm,
designed to estimate the parameters of probabilistic models from incomplete
datasets. The problem is as follows. If we had a complete database,
we would be able to estimate its parameters as explained in the previous
section. Alternatively, if we knew the parameters, we would be able
to generate a complete database by sampling from the probability distribution.

The EM algorithm proceeds by iteratively applying two steps. The E-step
(expectation) computes the probability $P(\mathbf{h}^{t}\,|\,\mathbf{v}^{t})$
for each configuration of the variables missing in~$\mathbf{v}^{t}$
in order to impute the missing values. More precisely, instead of
assigning a single value to each missing cell, we create a virtual
database in which all the configurations of~$\mathbf{H}^{t}$ are
present, each with probability $P(\mathbf{h}^{t}\,|\,\mathbf{v}^{t})$.
The M-step (maximization) uses this virtual complete database to adjust
the parameters of the model by MLE, as in Section~\ref{subsec:max-likelihood-selective}.
The two steps are repeated until the parameters (the weights) converge.

The problem is that initially we have neither a complete database
nor parameters for sampling the values of the missing variables. The
algorithm can be initialized by assigning arbitrary values to the
parameters or by assigning arbitrary values to the variables in~$\mathbf{Z}$.
Unfortunately, a bad choice of the initial values may cause the algorithm
to converge to a local maximum of the likelihood, which may be quite
different from the global maximum.

The~$n_{ij}$'s required by the M-step are obtained by counting the
number of instances in the database for which the link $(i,j)$ belongs
to the tree induced by $\mathbf{v}^{t}\mathbf{h}^{t}$. These are
the $n_{ij}$'s introduced in Equation~\ref{eq:nij-log-wij}, which
in the case of the virtual database are
\begin{equation}
n_{ij}=\sum_{t=1}^{T}\:\sum_{\mathbf{h}^{t}\,|\,(i,j)\in{\cal T}'_{\mathbf{v}^{t}\mathbf{h}^{t}}}P'(\mathbf{h}^{t}\,|\,\mathbf{v}^{t})\label{eq:nij-EM-1}
\end{equation}
and can be efficiently computed by applying the following result:
\begin{prop}
\label{prop:nij-EM}Given a database~$\mathcal{D}$ with~$T$ instances
and a selective SPN, the $n_{ij}$'s in Equation~\ref{eq:nij-EM-1}
are
\begin{equation}
n_{ij}=\sum_{t=1}^{T}w_{ij}\cdot S_{i}^{\partial}(\mathbf{v}^{t})\cdot S_{j}(\mathbf{v}^{t})\;.\label{eq:nij-EM-2}
\end{equation}
\end{prop}
Once we have the $n_{ij}$'s, the weights can be updated using Equation~\ref{eq:estimation-w_ij}
or~\ref{eq:estimation-w_ij-Laplace}. The time required by each iteration
of EM is proportional to the size of the network and the number of
instances in the dataset.

\paragraph{Hard EM}

The EM algorithm needs the value of $S_{i}^{\partial}$, which is
proportional to $\partial S/\partial w_{ij}$ and may thus be very
small when the link $(n_{i},n_{j})$ is in a deep position, i.e.,
far from the root. Therefore this algorithm may suffer from the vanishing
gradients problem in the same way as GD\@. To avoid it, Poon and
Domingos \cite{poon2011} proposed a \emph{hard} version of EM for
SPNs that selects for each hidden variable $H\in\mathbf{H}^{t}$ the
most probable state. Thus, in the E-step of each iteration, every
instance of the dataset contributes to the update of just one weight
per sum node, instead of contributing to all of them proportionally.

Hsu et\ al.~\cite{hsu2017} proposed a variant of hard EM for SPNs
with Gaussian leaves. This method proceeds top-down. It starts at
the top sum node and distributes the instances among its children
by maximum likelihood. Every other sum node receives only the instances
distributed to its parents and redistributes them in the same fashion.
This way it updates the weights of the sum nodes locally. The process
is similar to the automatic parameter learning in LearnSPN (cf.\ Sec.~\ref{subsec:LearnSPN}).
These authors also provide formulas for updating the parameters of
Gaussian leaves.

\subsubsection{Comparison of MLE algorithms\label{subsec:CCCP}}

The application of the EM to SPNs has been justified with different
mathematical arguments. Peharz \cite{peharz2015} exploited the interpretation
of the sum nodes in the augmented network as the sum of conditional
probability functions (cf.\ Eqs.~\ref{eq:sum-condic} and~\ref{eq:sum-condic-2}).
In turn, Zhao et~al.\ \cite{zhao2016a}, using a unified framework
based on signomial programming, designed two algorithms for learning
the parameters of SPNs: sequential monomial approximations (SMA) and
the concave-convex procedure (CCCP). GD is a special case of SMA,
while CCCP coincides with EM in the case of SPNs, despite being different
algorithms in general. Their experiments proved that EM/CCCP converges
much faster than the other algorithms, including GD\@. In turn, Desana
and Schnörr \cite{desana2017} derived the EM algorithm for SPNs whose
leaf nodes may represent complex probability distributions.

In discriminative learning, neither EM nor CCCP have a closed-form
expression for updating the weights \cite{gens2012}. Rashwan et~al.\ \cite{rashwan2018}
addressed this problem with the extended Baum-Welch (EBW) algorithm,
which updates the parameters of the network using a transformation
that increases the value of the likelihood function monotonically.
In the generative case, this transformation coincides with the update
formula of EM/CCCP (the M-step), while in the discriminative case
it\textbf{ }provides a method to maximize the (conditional) likelihood
function with a closed-form formula. They also adapted this method
to SPNs with Gaussian leaves.

Both the algorithm of Desana and Schnörr and EBW outperformed GD and
EM in a wide variety of datasets.

\subsection{Semi-supervised learning}

Trapp et\ al.~\cite{trapp2017} introduced a safe semi-supervised
learning algorithm for SPNs. By ``safe'' they mean that the model
performance can be increased but never degraded by adding unlabeled
data. They extended the EM to generative semi-supervised learning
and defined a discriminative semi-supervised learning approach. They
also introduced the maximum contrastive pessimistic algorithm (MCP-SPN),
based on \cite{loog2016}, for learning safe semi-supervised SPNs.
Their results were competitive with those of purely supervised algorithms.

\subsection{Approximate Bayesian learning \label{subsec:Approx-Bayesian}}

There are alternative methods for learning the parameters of an SPN
based on approximate Bayesian techniques, such as Bayesian moment
matching \cite{rashwan2016} and collapsed variational inference \cite{zhao2016b},
which are not as exposed to overfitting as GD or EM\@. Both Bayesian
methods start with a product of Dirichlet distributions as a prior;
the posterior distribution $P(w_{ij}\,|\,\mathcal{D})$ is a mixture
of products of Dirichlets, which is computationally intractable. In
both works the solution applied was to approximate that distribution
with a single product of Dirichlet distributions. Rashwan et\ al.~\cite{rashwan2016}
applied online Bayesian moment matching (oBMM), which approximates
the posterior distributions of the weights by computing a subset of
their moments and finding another distribution from a tractable family
that matches those moments. In this case, it sufficed to match the
first and second order moments of the distribution. The experiments
showed that this approach outperforms SGD and online EM. This method
has also been adapted to SPNs with Gaussian leaves by Jaini et\ al.~\cite{jaini2016}.
In the same vein, Zhao and Gordon \cite{zhao2017} presented an optimal
linear time algorithm for computing the moments in SPNs with general
acyclic directed graph structures, based on the mixture of trees interpretation
of SPNs. This provides an effective method to apply Bayesian moment
matching to a broad family of SPNs.

As mentioned above, Zhao et\ al.~\cite{zhao2016b} addressed the
problem by applying collapsed variational Bayesian inference (CVB-SPN).
This approach treats the dataset as partial evidence, whose missing
values correspond to the latent variables of the SPN\@. They assumed
that the missing values of each instance are not independent of those
missing in other instances, and marginalized these variables out of
the joint posterior distribution (the “collapse” step). Then, they
approximated this distribution with the product of the Dirichlets
that maximizes some evidence lower bound (ELBO) of the log-likelihood
function of the dataset (the “variational inference” step). The experiments
showed that the online version of CVB-SPN---i.e., the version that
receives a stream of data in real time---outperforms oBMM in many
datasets.

\subsection{Deep learning approach}

Peharz et\ al.~\cite{peharz2018} considered a special class of
SPNs, which they called random SPNs, and trained them with automatic
differentiation, stochastic GD, and dropout, using GPU parallelization.
The resulting model was called RAT-SPN\@. Its classification accuracy,
measured on the MNIST images and other databases, was comparable to
that of deep neural networks, with the advantages of a probabilistic
generative model, such as interpretability and robustness to missing
features. 

\section{Structural learning\label{sec:Structure-learning}}

Structural learning consists in finding the optimal (or near-optimal)
graph of an SPN\@. Most of the algorithms for this task require some
computation of probabilities during the process.

\subsection{First structure learners}

BuildSPN, by Dennis and Ventura \cite{dennis2012} was the first algorithm
of this kind. It looks for subsets of highly correlated variables
and introduces latent variables to account for those dependencies.
These variables generate sum nodes and the process is repeated recursively
looking for the new latent variables.

BuildSPN and the hand-coded structure of Poon and Domingos \cite{poon2011},
both designed for image processing, assumed neighborhood dependence.
In order to overcome that limitation, Peharz et\ al.~\cite{peharz2013}
proposed an algorithm that subsequently combines SPNs of few variables
into larger ones applying a statistical dependence test.

BuildSPN was also critiqued by Gens and Domingos \cite{gens2013}
because (1) the clustering process may separate highly dependent variables,
(2) the size of the SPN and the time required can grow exponentially
with the number of variables, and (3) it requires an additional step
to learn the weights.

\subsection{LearnSPN\label{subsec:LearnSPN}}

\begin{figure}
\begin{centering}
\includegraphics[scale=0.64]{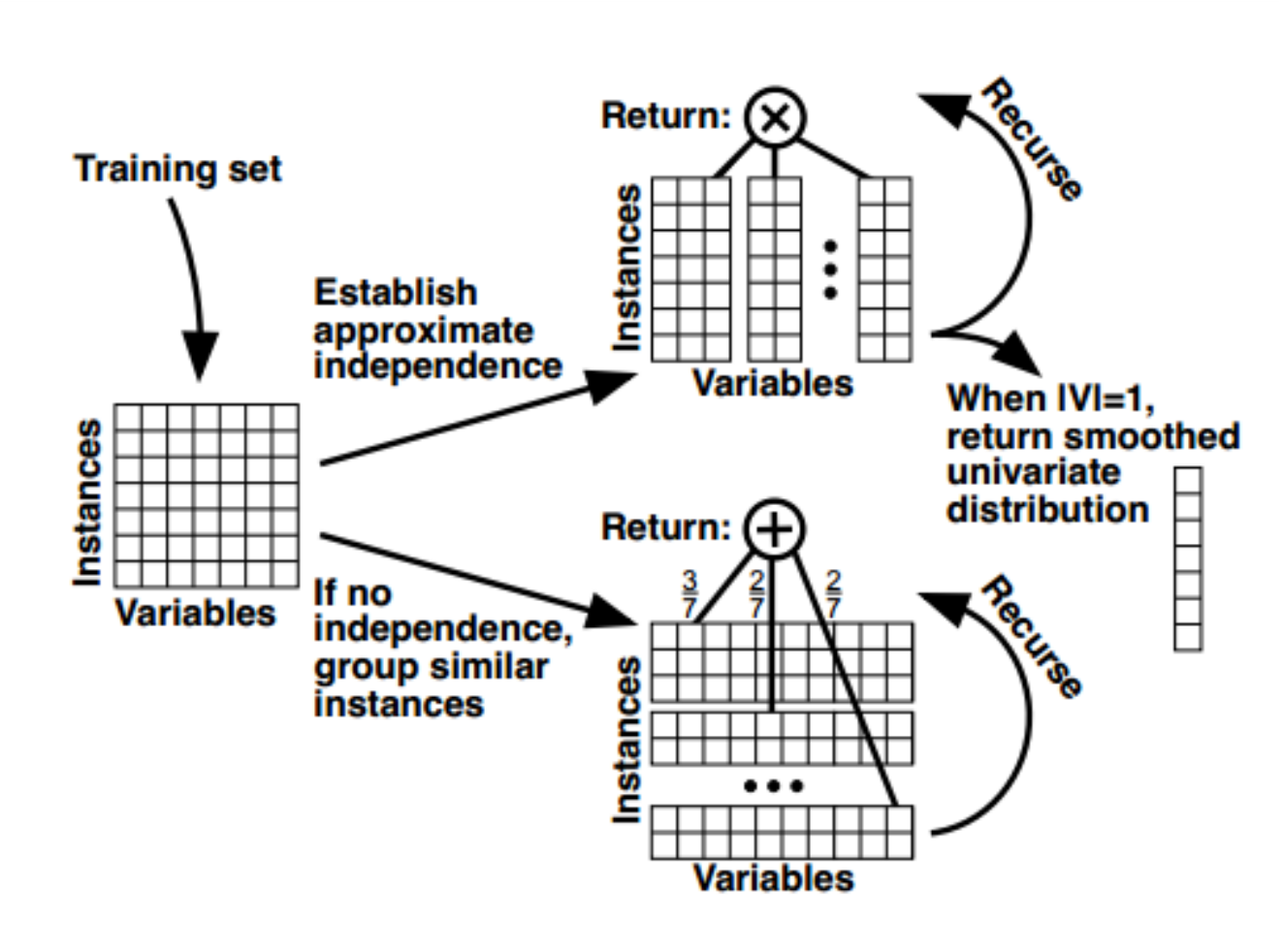}
\par\end{centering}
\caption{\label{fig:The-LearnSPN-algorithm}The LearnSPN algorithm recursively
creates a product node when there are subsets of (approximately) independent
variables and a sum node otherwise, grouping similar instances. (Reproduced
from \protect\cite{gens2013} with the authors' permission.)\label{fig:LearnSPN}}
\end{figure}

It is common in machine learning to see a dataset as a \emph{data
matrix} whose columns are attributes or variables and whose rows are
observations or instances. The LearnSPN algorithm \cite{gens2013}
recursively splits the variables into independent subsets (thus ``chopping''
the data matrix, as shown in Figure \ref{fig:The-LearnSPN-algorithm})
and then clusters the instances (thus ``slicing'' the matrix). Every
``chopping'' creates a product node and every ``slicing'' a sum
node, as indicated in Algorithm~\ref{alg:LearnSPN}. There are two
base cases:
\begin{enumerate}
\item When the piece of the data matrix produced by ``chopping'' contains
a single column (i.e., one variable), the algorithm creates a terminal
node with a univariate distribution using MLE.
\item When the piece of the data matrix produced by ``slicing'' contains
several columns with relatively few rows, the algorithm applies a
naïve Bayes factorization over those variables. This is like ``chopping''
that piece into individual columns, which will be processed as in
the base case~1.
\end{enumerate}

LearnSPN can be seen as a framework algorithm in the sense that it
does not specify the procedures for splitting independent subsets
of variables (\emph{splitVariables} in Algorithm~\ref{alg:LearnSPN})
and clustering similar instances (\emph{clusterInstances }in that
algorithm). Originally Gens and Domingos \cite{gens2013} chose the
G-Test for splitting and hard incremental EM for clustering.

\begin{algorithm}
\caption{\label{alg:LearnSPN}$\text{\textsf{LearnSPN}}(\mathbf{T},\mathbf{V},\alpha,m)$}

\textbf{~~Input:} $\mathbf{T}$: a data matrix of instances over
the variables in~$\mathbf{V}$; $m$: minimum number of instances
to allow a split of variables; $\alpha$: Laplace smoothing parameter

\textbf{~~Output:} an SPN $\mathcal{S}$ with $\text{sc}({\cal S})=\mathbf{V}$

\textbf{~~if $|\mathbf{V}|=1$ then}

~~~~~~$\mathcal{S}\leftarrow\text{ \textsf{univariateDistribution}}(\mathbf{T},\mathbf{V},\alpha)$

\textbf{~~else if} $|\mathbf{T}|<m$ \textbf{then}

~~~~~~$\mathcal{S}\leftarrow\text{\textsf{naïveFactorization}}(\mathbf{T},\mathbf{V},\alpha)$

\textbf{~~else}

~~~~~~$\left\{ V_{j}\right\} _{j=1}^{C}\leftarrow\text{\textsf{splitVariables}}(\mathbf{T},\mathbf{V},\alpha)$

~~~~~~\textbf{if} $C>1$ \textbf{then}

~~~~~~~~~~$\mathcal{S}\leftarrow\prod_{j=1}^{C}\text{\textsf{LearnSPN}}\left(\mathbf{T},\mathbf{V}_{j},\alpha,m\right)$

~~~~~~\textbf{else}

~~~~~~\textbf{~~~~}$\left\{ \mathbf{T}_{i}\right\} _{i=1}^{R}\leftarrow\text{\textsf{clusterInstances}}(\mathbf{T},\mathbf{V})$

~~~~~~\textbf{~~~~}$\mathcal{S}\leftarrow\sum_{i=1}^{R}\frac{\left|\mathbf{T}_{i}\right|}{|\mathbf{T}|}\text{\textsf{LearnSPN}}\left(\mathbf{T}_{i},\mathbf{V},\alpha,m\right)$

\textbf{~~return} $\mathcal{S}$
\end{algorithm}

Splitting the variables (``chopping'') only considers pair-wise
independencies. The process starts with a graph containing a node
for each variable and no links. It randomly selects one variable and
adds an edge to the first other variable deemed dependent by the G-test,
then moves to that variable, and iterates until no new variable can
be linked to this component of the graph. At the end, if this component
has gathered all variables only one component is generated; then the
clustering concludes and the algorithm clusters instances instead.

Clustering similar instances (``slicing'') is achieved by the hard
EM algorithm assuming a naïve Bayes mixture model, where the variables
are independent given the cluster $\mathbf{C}_{i}$: $\ensuremath{P(\mathbf{v})=\sum_{i}P\left(\mathbf{c}_{i}\right)\prod_{j}P\left(v_{j}\,|\,\mathbf{c}_{i}\right)}$.
This particular model produces a clustering that can be chopped in
the next recursion. This version of LearnSPN forces a clustering in
the first step, without attempting a split.

\subsection{ID-SPN}

Rooshenas and Lowd \cite{rooshenas2014} observed that PGM learners
usually analyze direct interactions (dependencies) between variables
while previous SPN learners analyze indirect interactions (dependencies
through a latent variable). The indirect-direct SPN (ID-SPN) structure
learner combines both methods. Their initial idea is that any tractable
multivariate distribution that can be represented as an arithmetic
circuit or an SPN can be the leaf of an SPN without losing tractability.
With this idea they learned arithmetic circuit Markov networks (ACMN)
\cite{lowd2013}, which are roughly Markov networks learned as arithmetic
circuits. ID-SPN begins with a singular ACMN node and tries to replace
it with a mixture (yielding a sum node) or a product (yielding a product
node), similar to the\emph{ }cluster\emph{ }and split operations in
LearnSPN\@. If a replacement increases the likelihood, it is saved
and the algorithm recurs on the new ACMN leaves, until the likelihood
does not increase. This top-down process represents the learning of
indirect interactions, while the creation of ACMN leaves represents
the learning of direct interactions. This algorithm outperformed all
previous algorithms and is currently the state of the art. However,
ID-SPN is slower and more complex than LearnSPN, and has many more
hyperparameters to tune, which requires a random search in the space
of hyperparameters instead of a grid search.

\subsection{Other algorithms}

Peharz et\ al.~\cite{peharz2014a} proposed a structure learner
that searches within the space of selective SPNs and showed that it
is competitive with LearnSPN.

Adel et\ al.~\cite{adel2015} pointed out that previous work had
only compared algorithms on binary datasets. They designed SVD-SPN,
which proceeds by finding rank-1 matrices. This allows the algorithm
to cluster and split at the same time, producing optimal data matrix
pieces. It operates recursively, like LearnSPN, but constructs the
SPN from the rank-1 submatrices extracted. It also considers a multivariate
base case when the variables in the pieces of the data matrix are
highly correlated. In this case a sum node is created with as many
children as instances in the piece of the matrix; each child is a
product node of all the variables in the matrix. In their experiments,
SVD-SPN obtained results similar to those of LearnSPN and ID-SPN for
binary datasets, but outperformed them in multiple-category datasets,
such as Caltech-101, and is~5 times faster.

\subsection{\label{sec:Improvements-to-LearnSPN}Improvements to LearnSPN}

Even though LearnSPN is not the best performing algorithm, it is still
widely used owing to its simplicity and modularity \cite{butz2018b}
and has led to several variants.

\subsubsection{Algorithm of Vergari et\ al.~\cite{vergari2015}}

\begin{figure}
\begin{centering}
\includegraphics[scale=0.22]{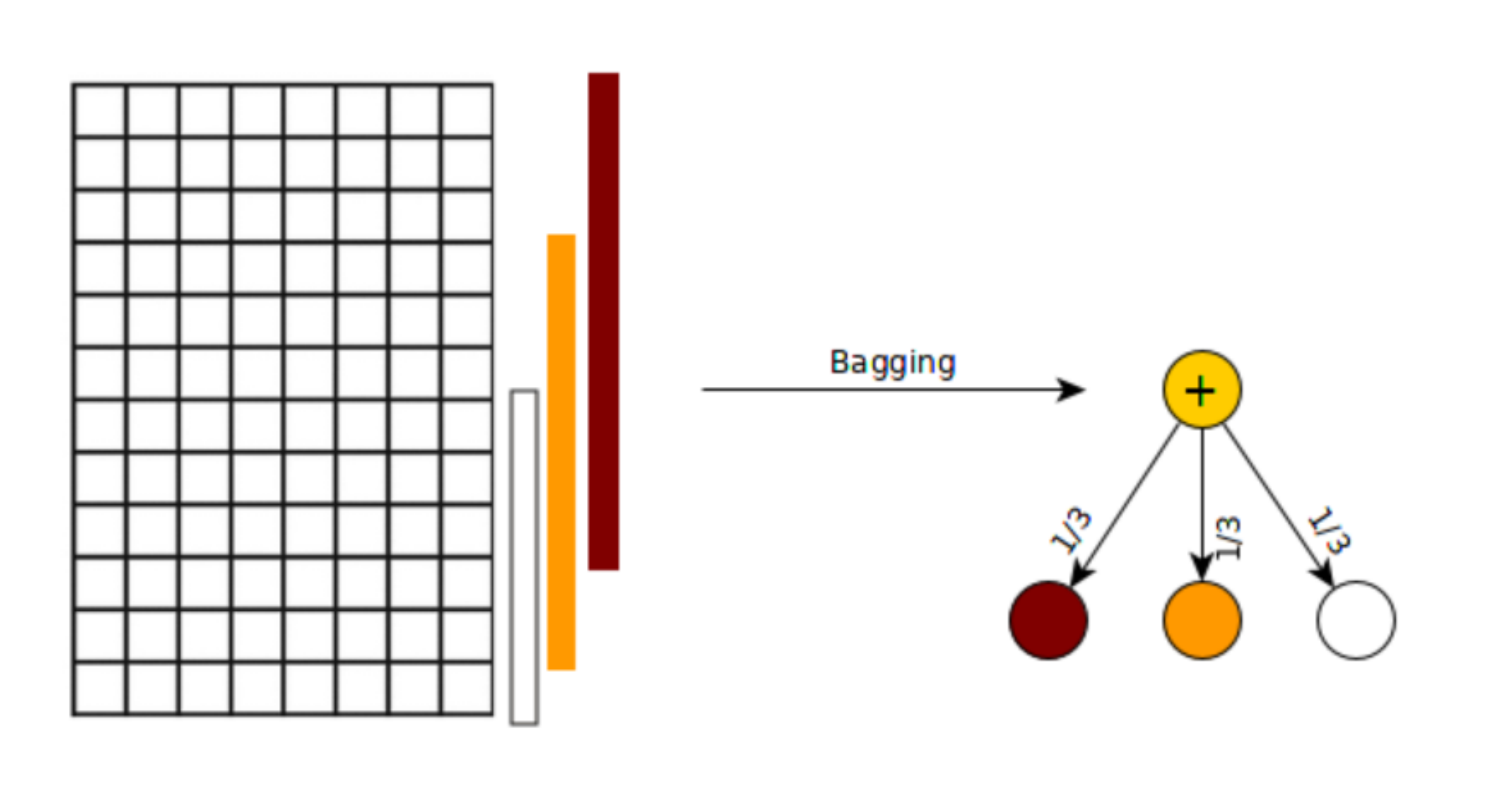}
\par\end{centering}
\caption{\label{fig:Bagging-in-SPNs.}Learning SPNs with bagging.}
\end{figure}

It consists of three modifications to LearnSPN:
\begin{enumerate}
\item \emph{Binary splits}. Every split cuts the data matrix into only two
pieces. This avoids creating too complex structures at early stages
(which often occurs when learning from noisy data) and favors deep
structures over shallow ones. This is not a limitation in the number
of children of product nodes because consecutive splits can be applied
if necessary.
\item \emph{Chow-Liu trees (CLTs) in the leaves}. The naïve Bayes factorization
used as the base case of LearnSPN (see Algorithm~\ref{alg:LearnSPN})
can be replaced by the creation of Chow-Liu trees \cite{chow1968},
which are equivalent to tree-shaped Bayesian networks or Markov networks.
Every tree is built by linking the variables with higher mutual information
until there is a path between every pair of variables. CLTs are more
expressive than the naïve Bayes factorization (which is a particular
case of CLT) without adding computational complexity. LearnSPN stops
earlier when using CLTs as leaves because each tree can accommodate
more instances, thus yielding simpler SPN structures (with fewer edges)
with lower risk of overfitting.
\item \emph{Bagging.} This technique, originally used to build random forests
\cite{breiman2001}, consists in taking several random samples from
a dataset, each consisting of several instances, and building a classifier
for each sample. The overall classification can be the average of
the outputs of the individual classifiers (for continuous variables)
or the mode (for finite-state variables). In SPN learning, it extracts---with
replacement---$n$~samples of the dataset and produces a sum node
with $n$~children, setting every weight to $1/n$, as shown in Figure~\ref{fig:Bagging-in-SPNs.}.
Each child represents one of the individual classifiers and the sum
node averages the $n$~results. Since the network size would grow
exponentially if bagging were applied before every clustering, it
is applied only before the first LearnSPN operation---which is a
clustering---in order to achieve the widest effect on the resulting
structure.
\end{enumerate}

The experiments showed that (1) binary splits yield deeper and simpler
SPNs and generally reduce the number of edges and parameters, (2)
using Chow-Liu trees attains the same effect and generally increases
the likelihood, and (3) bagging also increases the likelihood, especially
in datasets with a low number of instances. With these modifications,
LearnSPN achieved the same performance as ID-SPN.

\subsubsection{Beyond tree SPNs}

One of the main disadvantages of both LearnSPN and ID-SPN is that
they always produce trees (except when the leaves are Markov networks).
In order to generate more compact SPNs, Dennis and Ventura \cite{dennis2015}
designed SearchSPN, an algorithm that produces SPNs in which nodes
may have several parents. It selects the product node that contributes
less to the likelihood and greedily searches for candidate structures
using modified versions of the clustering methods of LearnSPN\@.
The resulting likelihood is significantly better than that of LearnSPN
for the majority of datasets and comparable with that of ID-SPN, but
on average the execution is 7 times faster and the number of nodes
10 times smaller.

In the same vein, Rahman and Gogate \cite{rahman2016} created a post-processing
algorithm that, after applying LearnSPN with CLTs in the leaves, merges
similar sub-SPNs. Similarity is measured with a Manhattan distance;
if two sub-SPNs are closer than a certain threshold, the pieces of
the data matrix from which they come are combined and the algorithm
chooses the sub-SPN with the higher likelihood for the combined data.
This modification of LearnSPN increases the likelihood and reduces
the number of parameters of the SPN; additionally, it dramatically
increases the learning time for some datasets. In combination with
bagging, it outperformed other algorithms---including ID-SPN \cite{rooshenas2014}---for
high-dimensional datasets.

\subsubsection{Further improvements to LearnSPN}

As mentioned in Section~\ref{subsec:CCCP}, Zhao et\ al.~\cite{zhao2016a}
showed that learning the parameters with the CCCP algorithm improves
the performance of LearnSPN.

Di Mauro et\ al.~\cite{dimauro2017} proposed approximate splitting
methods to accelerate LearnSPN, thus trading speed for quality (likelihood).

Butz et\ al.~\cite{butz2018b} studied the different combinations
of algorithms for LearnSPN. They compared mutual information and the
G-test for splitting, and $k$-means and Gaussian mixture models for
clustering. The best results were obtained when using the G-test and
either $k$-means or Gaussian mixture models, both for the standard
LearnSPN and for the version that generates CLTs in the leaves.

Liu et\ al.~\cite{liu2019} proposed a clustering method that decides
the number of instance clusters adaptively, i.e., depending on each
piece of data matrix evaluated. Their goal was to generate more expressive
SPNs, in particular deeper ones with controlled widths. When compared
previous algorithms (namely, standard LearnSPN \cite{gens2013}, LearnSPN
with binary splits \cite{vergari2015}, and LearnSPN with approximate
splitting \cite{dimauro2017}), their method achieved higher likelihood
in 20 binary datasets and generated deeper networks (i.e., more expressive
SPNs) while maintaining a reasonable size.

\subsubsection{LearnSPN with piecewise polynomial distributions\label{subsec:learn-multivariate}}

Most learning algorithms assume that when a terminal node represents
a continuous variable, the univariate distribution belongs to a known
family (Poisson, Gaussian, etc.) and only the parameters must be optimized.
However, there are at least two variants of LearnSPN that, in addition
to having indicators for finite-state variables, use a piecewise polynomial
distribution for each leaf node representing a numeric variable, instead
of requiring the user to specify a parametric family \cite{bueff2018,molina2018}.

In LearnWMISPN \cite{bueff2018}, which combines LearnSPN with weighted
model integration (WMI), the order of each polynomial is determined
using the Bayesian information criterion (BIC) \cite{schwarz1978}.
A preprocessing step transforms finite-state, categorical, and continuous
features into a binary representation before applying LearnSPN\@.
The corresponding inference algorithm can answer complex conditional
queries involving both intervals for continuous variables and values
for discrete variables.

In mixed SPN (MSPNs) \cite{molina2018} the operations of decomposition
(splitting) and conditioning (clustering) are based on the Hirschfeld-Gebelein-Rényi
maximum correlation coefficient \cite{gebelein1941}.

\subsection{Online structural learning}

The algorithms presented so far need the complete dataset to produce
a structure. However, sometimes the dataset is so big that the computer
does not have enough memory to store it at once. In other situations,
e.g., in recommender systems, the data arrive constantly. In these
cases the learning algorithm must be able to update the structure
instead of learning it from scratch every time new data arrives.

In this context, Lee et\ al.~\cite{lee2013} designed a version
of LearnSPN where clustering (slicing) is replaced by online clustering,
so that new sum children can be added when new data arrive, while
product nodes are unmodified.

Later Dennis and Ventura \cite{dennis2017} extended their SearchSPN
algorithm \cite{dennis2015} to the online setting. This online version
is as fast as the offline version that works only on the current batch
and the quality of the resulting SPN is the same.

Hsu et\ al.~\cite{hsu2017} created oSLRAU, an online structure
learner for Gaussian leaves (oSLRAU) which begins with a completely
uncorrelated SPN structure that is updated when the arriving data
reveals a new correlation. The update consists in replacing a leaf
with a multivariate Gaussian leaf or a mixture over its scope.

Jaini et\ al.~\cite{jaini2018} proposed an algorithm, Prometheus,
whose first concern is to avoid the parameter that decides when two
subsets of variables are independent in order to perform a LearnSPN
split. So instead of creating a product node, it creates a mixture
of them, representing different subset partitions. The way the partitions
are created allows them to share subsets, which is reflected in the
structure by common children, thus overcoming the restriction to trees
on the way. This is in some sense similar to bagging in sum nodes
(cf.\ Sec.~\ref{sec:Improvements-to-LearnSPN}) and makes the algorithm
robust in low data regimes. However, the complexity of the algorithm
grows with the square of the number of variables. In order to extend
it to high-dimensional datasets, the authors created a version that
samples in each step from the set of variables instead of using all
of them. This algorithm can treat discrete, continuous, and mixed
datasets. Their experiments showed that this algorithm surpasses both
LearnSPN and ID-SPN in the three types of datasets. It is also robust
in low data regimes, achieving the same performance as oSLRAU with
only 30-40\% of the data.

\subsection{Learning with dynamic data}

Data are said to be \emph{dynamic} when all the variables (or at least
some of them) have different values in different time points---for
example, \emph{Income-at-year-1}, \emph{Income-at-year-2}, etc. The
set of variables for a specific time point is usually called a \emph{slice}.
The slice structure, called \emph{template, }is replicated and chained
to accommodate as many time points as necessary. The length of the
chain is called the \emph{horizon}.

For this problem Melibari et\ al.~\cite{melibari2016} proposed
dynamic SPNs (DSPNs), which extend SPNs in the same way that dynamic
Bayesian networks (DBNs) \cite{dean1989} extend BNs. A local-search
structure learner generates an initial template SPN and searches for
neighboring structures, trying to maximize the likelihood. Every neighbor,
which comes from replacing a product node, represents a specific choice
of factorization of the variables in its scope. The algorithm searches
over other choices of factorizations and updates the structure if
a better one is found. This method outperforms non-dynamic algorithms,
such as LearnSPN, and other models, such as dynamic Bayesian networks
and recurrent neural networks.

Later, Kalra et\ al.~\cite{kalra2018} extended oSLRAU to the dynamic
setting by unrolling the SPN to match the length of the chain to the
horizon, with shared weights and a shared covariance matrix, to decide
when a new correlation requires a change in the template. This algorithm
surpassed that of Melibari et\ al.~\cite{melibari2016} and hidden
Markov models in 5 sequential datasets, and recurrent neural networks
in~4 of those datasets.

\subsection{Relational data learning\label{subsec:relational-SPNs}}

Nath and Domingos \cite{nath2015} introduced relational SPNs (RSPNs),
which generalize SPNs by modeling a set of instances jointly, allowing
them to influence each other’s probability distributions, and modeling
probabilities of relations between objects. Their LearnRSPN algorithm
outperformed Markov Logic Networks in both running time and predictive
accuracy when tested on three datasets.

\subsection{Bayesian structure learning}

Lee et\ al.~\cite{lee2014} designed a Bayesian non-parametric extension
of SPNs. Trapp et\ al.~\cite{trapp2016} criticized this work for
neglecting induced trees in their posterior construction, corrected
it by introducing \emph{infinite sum-product tree},\emph{ }and showed
that it yields higher likelihood than infinite Gaussian mixture models.

A common problem of structural learning algorithms is the lack of
a principled criterion for deciding what a ``good'' structure is.
For this reason, Trapp et\ al.~\cite{trapp2019} proposed an alternative
Bayesian approach that decomposes the problem into two phases: first
finding a graph and then learning the scope-function, $\psi$, which
assigns a scope to each node. The function~$\psi$ and the parameters
of the model are learned jointly using Gibbs sampling. The Bayesian
nature of this approach reduces the risk of overfitting, avoids the
need for a separate validation set to adjust the hyperparameters of
the algorithm, and enables robust learning of SPN structures under
missing data.

\section{Applications\label{sec:Applications}}

SPNs have been used for a wide variety of applications, from toy problems
to real-world challenges.

\subsection{Image processing}

\subsubsection{Image reconstruction and classification}

Poon and Domingos, in their seminal paper about SPNs \cite{poon2011},
applied them to image reconstruction, using a hand-designed structure
that took into account the local structure of the image data. They
tested their method on the datasets Caltech-101 and Olivetti. Then
Gens and Domingos \cite{gens2012} used a different hand-made structure
for image classification on the datasets CIFAR-10 and STL-10, obtaining
excellent results for that time, as mentioned in Section~\ref{subsec:SPNs-vs-PGMs}.

\subsubsection{Image segmentation}

Image segmentation consists in labeling every pixel with the object
it belongs to. Yuan et~al.\ \cite{yuan2016} developed an algorithm
that scales down every image recursively to different sizes and generates
object tags and unary potentials for every scale. Then, it builds
a multi-stacked SPN where every stack has a bottom and a top SPN\@.
The bottom SPN works on a pixel and its vicinity, going from the pixel
to bigger patches. Product nodes model correlations between patches
while sum nodes combine them into a feature of a bigger patch. When
the patch is as big as the pixel in the next scaled image, the results
are introduced in the top SPN alongside the unary potentials and the
tags of that scale. This process is stacked until the ``patch''
treated is the whole image. Multi-stacked SPNs have been especially
effective for handling occlusions in scenes.

Rathke et~al.\ \cite{rathke2017} applied SPNs to segmentation OCT
scans of retinal tissue. They first built a segmentation model for
the health model and for every pathology and then added to those models
typical shape variations of the retina tissue for some pathology-specific
regions. The resulting SPN extracts candidate regions (either healthy
or unhealthy) and selects the combination that maximizes the likelihood.
After a smoothing step, they obtain a complete segmentation of the
retina tissue as well as the diagnosis and the affected regions. This
method achieved state-of-the-art performance without needing images
labeled by pathologies.

\subsubsection{Activity recognition}

Wang and Wang \cite{wang2016} addressed activity recognition on still
images. They used unsupervised learning and a convolutional neural
network to isolate parts of the images, such as a hand or a glass,
and designed a spatial SPN including the spatial indicator nodes ``above'',
``below'', ``left'', and ``right'' for the product nodes to
encode spatial relations between pairs of these parts. They first
partitioned the image to consider only local part configurations.
Its SPN structure has two components: the top layers represent a partitioning
of the image into sub-images where product nodes act as partitions
and sum nodes as combinations of different partitions, while the bottom
layers represent the parts included in each sub-image and their relative
position using the spatial indicator nodes. In this sense the SPN
first learns spatial relations of isolated parts in sub-images and
then learns correlations between sub-images. Spatial SPNs outperformed
other activity recognition algorithms and were able to discover discriminant
pairs of parts for each class.

Amer and Todorovic \cite{amer2016} worked on activity localization
and recognition in videos. In their work, a \emph{visual word} is
a meaningful piece of an image, previously extracted with a neural
network. Visual words lie in a grid with three dimensions: height,
width, and time. Every grid position has a histogram of associated
visual words, called \emph{bag of words}. To construct an SPN, each
bag of words is treated as a variable with two states: foreground
and background. Product nodes represent a combination of sub-activities
into a more complex activity (for example, ``join hands + separate
hands = clap'') and sum nodes represent variations of the same activity.
An SPN is trained for every activity in a supervised context, in which
the foreground and the background values are known, and in a weakly
supervised context, in which only the activity is known. The structure
is a near-completely connected graph, pruned after parameter learning,
which proceeds iteratively: it learns---with GD---the weights of
the SPN from the parameters of the bag of words and then learns---with
variational methods---the parameters of the bag of words from the
weights of the SPN. The accuracy of this weakly supervised setting
was only 1.6 to 3\% worse than that of the supervised setting. This
approach in general achieved better performance than state-of-the-art
algorithms on several action-recognition datasets.

\subsubsection{Object detection}

Stelzner et\ al.~\cite{stelzner2019} proved that the attend-infer-repeat
(AIR) framework used for object detection and location is much more
efficient when the variational autoencoders (VAEs) that model individual
objects are replaced by SPNs: they achieved an improvement in speed
of an order of magnitude, with slightly higher accuracy, as well as
robustness against noise.

\subsection{Robotics}

Sguerra and Cozman \cite{sguerra2016} used SPNs for \emph{aerial
robots navigation}. Micro aerial vehicles need a set of sensors that
must comply with two criteria: light weight and real-time response.
Optical recognition with cameras satisfies the former while fast inference
with SPNs ensures the latter, as they were able to classify---in
real time---what the camera sees into pilot commands, such as ``turn
right'', and obtaining 75\% of accuracy with just 66 images.

Pronobis et\ al.~\cite{pronobis2017a} designed a probabilistic
representation of spatial knowledge called ``deep spatial affordance
hierarchy'' (DASH), which encodes several levels of abstractions
using a deep model of spatial concepts. It models knowledge gaps and
affordances by a deep generative spatial model (DGSM) which uses SPNs
for inference across different levels of abstractions. SPNs fit naturally
with DGSM because latent variables of the former are internal descriptors
in the latter. The authors tested it in a robot equipped with a laser-range
sensor.

Zheng et\ al.~\cite{zheng2017} designed graph-structured SPNs (GraphSPNs)
for structured prediction. Their algorithm learns template SPNs and
makes a mixture over them (a template distribution), which can be
applied to graphs of varying size re-using the same templates. The
authors applied them to model large-scale global semantic maps of
office environments with a exploring robot, obtaining better results
than with the classical approach based on undirected graphical models
(Markov networks).

The authors joined both models into an end-to-end deep model for semantic
mapping in large-scale environments with multiple levels of abstraction,
called TopoNets \cite{zheng2018}. These can perform inference about
unknown spatial information, are useful for novelty detection, and
achieve real-time performance.

\subsection{NLP and sequence data analysis}

Peharz et~al.\ \cite{peharz2014} applied SPNs to \emph{modeling
speech} by retrieving the lost frequencies of telephone communications
(artificial bandwidth extension). In this problem tractable and quick
(real-time) inference is essential. They used a hidden Markov model
(HMM) to represent the temporal evolution of the log-spectrum, clustered
the data using the Linde--Buzo--Gray algorithm and trained an SPN
for each cluster. The SPNs model each cluster and can be used to retrieve
the lost frequencies by MPE inference. This model has achieved better
results than state-of-the-art algorithms both objectively, with a
measure of log-spectral distortion, and subjectively, through listening
tests.

In language modeling, Cheng et\ al.~\cite{cheng2014} used a discriminative
SPN \cite{gens2012}, whose leaf nodes represent vectors with information
about previous words. This SPN was able to compute the probability
of the next word more accurately than classic methods for language
modeling, such as feedforward neural networks and recurrent neural
networks.

Later, Melibari et~al.\ \cite{melibari2016} used dynamic SPNs (DSPNs)
to analyze different sequence datasets. Unlike dynamic Bayesian networks,
for which inference is generally exponential in the number of variables
per time slice, inference in DSPNs has linear complexity. In a comparative
study with five other methods, including HMMs and neural networks
with long short-term memory (LSTM), DSPNs were superior in~4 of the~5
datasets examined.

Ratajczak et\ al.~\cite{ratajczak2018} replaced the local factors
of higher-order linear-chain conditional random fields (HO-LC-CRFs)
and maximum entropy Markov models (MEMMs) with SPNs. These outperformed
other state-of-the-art methods in optical character recognition and
achieved competitive results in phoneme classification and handwriting
recognition.

\subsection{Other applications}

Vergari et\ al.~\cite{vergari2018} used SPNs as autoencoders (SPAEs)
for feature extraction. They trained the SPNs with LearnSPN and used
the values of the internal nodes or the states of the latent variables
associated to sum nodes as the codification variables. Although this
model was not trained to reconstruct its inputs, experiments showed
that SPAEs are competitive with state-of-the-art autoencoder architectures
for several multilabel classification problems.

Butz et~al.\ \cite{butz2018a}\emph{ }used Bayesian networks to
recognize independencies in 3,500 datasets of \emph{soil bacteria}
and combined them into an SPN in order to efficiently compute conditional
probabilities and the MPE\@.

Nath and Domingos \cite{nath2016} used relational SPNs (cf.\ Sec.~\ref{subsec:relational-SPNs})
for \emph{fault localization}, i.e. finding the most probable location
of bugs in computer source code. The networks, trained on a corpus
of previously diagnosed buggy programs, learned to identify recurring
patterns of bugs. They could also receive clues about bug suspicion
from other bug detectors, such as \textsc{Tarantula}.

Hilprecht et\ al.~\cite{hilprecht2019} proposed learning database
management systems from data instead of queries, using ensembles of
relational SPNs. This approach provides better accuracy and better
generalization to unseen queries than different state of the art methods.

Vergari et\ al.~\cite{vergari2019a} also evaluated SPNs for representation
learning \cite{bengio2013}. SPNs encode a hierarchy of part-based
representations which can be ordered by scope length. When compared
with other representation learners, such as VAEs or random Boltzmann
machines (RBMs), they provided competitive results both in supervised
and semi-supervised settings. Moreover, the model trained for extracting
representations can be used as-is for inference.

Vergari et\ al.~\cite{vergari2019b} designed a tool for automatic
exploratory data analysis without the need for expert statistical
knowledge. It leverages Gibbs sampling and a modification of mixed
SPNs \cite{molina2018} to model the data, and provides functionalities
such as data type recognition, missing values imputation, anomaly
detection, and others.

Roy et\ al.~\cite{roy2019} addressed the explanation of activity
recognition and localization in videos. A deep convolutional neural
network is used for localization and its output is introduced to an
SPN. Both models are learned jointly. The explainability of the system
was evaluated by the subjective user trust in the explanations that
the SPN provides about the criteria of the neural net.

\section{Software for SPNs\label{sec:Software}}

Every publication about SPNs presents some experiments, and in many
cases the source code is publicly available. The web page \url{https://github.com/arranger1044/awesome-spn}
contains many references about SPNs, classified by year and by topic;
the section ``Resources'' includes links to talks and tutorials,
the source code for some of those publications, and several datasets
commonly used for the experiments. Most of the software is written
in Python or C++.

In particular, there are two projects that aim to develop comprehensive,
simple, and extensible libraries for SPNs. Both of them are written
in Python and use TensorFlow as a backend for speeding up some operations.
\textsf{LibSPN},\footnote{\href{https://www.libspn.org}{https://www.libspn.org}.}
initiated by Andrzej Pronobis at the University of Washington, Seattle,
WA \cite{pronobis2017b}, implements several methods for inference
(marginal and conditional probabilities, and approximate MPE), parameter
learning (batch and online, with GD and hard EM), and visualization
of SPNs. It lacks algorithms for structural learning, but it allows
building convolutional SPNs with a layer-oriented interface \cite{wolfshaar2019}.
The SPNs, stored as Python structures, are compiled into TensorFlow
graphs for parameter learning and inference; for this purpose \textsf{LibSPN}
has implemented in C++ and CUDA some operations that cannot be performed
efficiently with native TensorFlow operations. Several tutorials in
Jupyter Notebook are available at its website. It has been used mainly
for computer vision and robotics \cite{pronobis2017a,zheng2017,wolfshaar2019}.

The other library, \textsf{SPFlow}\footnote{\href{https://github.com/SPFlow/SPFlow}{https://github.com/SPFlow/SPFlow}.},
is developed by Alejandro Molina at the University of Darmstadt, Germany,
with contributors from several countries \cite{molina2019}. It implements
methods for inference (marginal and conditional probabilities, and
approximate MPE), parameter learning (with GD) and several structural
learning algorithms, and can be extended and customized to implement
new algorithms. SPNs are usually compiled into TensorFlow for fast
computation, but they can also be compiled into C, CUDA, or FPGA code.

There are also some smaller libraries of interest, such as \textsf{SumProductNetworks.jl}\emph{
}for Julia,\footnote{\href{https://github.com/trappmartin/SumProductNetworks.jl}{https://github.com/trappmartin/SumProductNetworks.jl}.}
which implements inference and parameter learning, and the \textsf{Libra
Toolkit} \cite{lowd2015},\footnote{\href{http://libra.cs.uoregon.edu}{http://libra.cs.uoregon.edu}.}
a collection of algorithms written in OCaml for learning several types
of probabilistic models, such as BNs, SPNs, and others, including
the ID-SPN algorithm \cite{rooshenas2014}.

\section{Extensions of SPNs\label{sec:Extensions}}

In recent years there have been several extensions of SPNs to more
general models. In this section we briefly comment on some of them.

Sum-product-max networks (SPMNs) \cite{melibari2016a} generalize
SPNs to the class of decision making problems by including two new
types of nodes, max and utility, like in influence diagrams. These
networks can compute the expected utility and the optimal policy in
time proportional to the number of links.

Autoencoder SPNs (AESPNs) \cite{dennis2017b} combine two SPNs with
an autoencoder between them. This model produces better samples than
SPNs by themselves.

Tan and Peharz \cite{tan2019} designed a mixture model of VAE pieces
over local subsets of variables combined via an SPN\@. This combination
yields better density estimates, smaller models and improved data
efficiency with respect to VAEs.

In credal sum-product networks (CSPNs) \cite{maua2017} the weights
of each sum node do not have a fixed value, but vary inside a set
(a product of probability simplexes) in such a way that each choice
of the weights defines an SPN.

Sum-product graphical models (SPGMs) \cite{desana2020} join the semantics
of probabilistic graphical models with the evaluation efficiency and
expressiveness of SPNs by allowing the nodes associated to variables
to appear in any part of the network, not only in the leaf nodes.
Their LearnSPGM algorithm outperformed both LearnSPN and ID-SPN on
the 20 real-world datasets previously used in \cite{lowd2008}.

Sum-product-quotient networks (SPQNs) \cite{sharir2018a} introduce
quotient nodes, which take two inputs and output their quotient, allowing
these models to represent conditional probabilities explicitly.

Tensor SPNs (tSPNs) \cite{ko2018} are an alternative representation
of SPNs. Their main advantage is an important reduction in the number
of parameters---between 24 and 569 times in the experiments---with
little loss of modeling accuracy. Additionally, tSPNs allow for faster
inference and a deeper and more narrow neural-network architecture.

Convolutional SPNs (ConvSPNs or CSPN) were developed independently
in \cite{wolfshaar2019} and \cite{butz2019}. The first model exploits
the inherent structure of spatial data in a similar way to convolutional
neural networks by using the sum and product operations of SPNs. The
second one finds a criterion for seeing when a convolutional neural
network is an SPN.

Submodular SPNs (SSPNs) \cite{friesen2017a} are an extension of SPNs
for scene understanding, whose weights can be defined by submodular
energy functions. 

Compositional kernel machines (CKMs) \cite{gens2017} are an instance-based
method closely related to SPNs. They have been successfully applied
to image processing tasks, mainly to object recognition.

Conditional SPNs (CSPNs) \cite{shao2019} extend SPNs to conditional
probability distributions. They include a new type of node, called
a \emph{gating node}, which computes a convex combination of the conditional
probability of its children with non-fixed weights.

The tutorial by Vergari, Choi, Peharz and Van den Broek at AAAI-2020
offers an excellent review of \emph{probabilistic circuits}, including
arithmetic circuits, SPNs, cutset networks (CNets), probabilistic
sentential decision diagrams (PSDDs), with many references for inference,
learning, applications, hardware implementations, etc.\footnote{\href{http://starai.cs.ucla.edu/slides/AAAI20.pdf}{http://starai.cs.ucla.edu/slides/AAAI20.pdf}.}

\section{Conclusion\label{sec:Conclusion}}

SPNs are closely related to probabilistic graphical models (PGMs),
such as Bayesian networks and Markov networks, but have the advantage
of allowing the construction of tractable models from data. SPNs have
been applied to the same tasks as neural networks, mainly image and
natural language processing, which exceeded by far the capabilities
of PGMs. Even though deep neural networks yield in general better
results, SPNs have the possibility of automatically building the
structure from data and learning the parameters with either gradient
descent or some of the algorithms developed for probabilistic models.

In this paper we have tried to offer a gentle introduction to SPNs,
collecting information that is spread in many publications and presenting
it in a coherent framework, trying to keep the mathematical complexity
to the minimum necessary for describing with rigor the main properties
of SPNs, from their definition to the algorithms for parametric and
structural learning. We have intentionally avoided presenting SPNs
as representations of network polynomials; readers interested in them
can consult \cite{peharz2015} and references therein. We have then
reviewed several applications of SPNs in different domains, some extensions,
and the main software libraries for SPNs. Given the rapid growth of
the literature about SPNs, some sections of this paper might become
obsolete soon, but we still hope it will be useful for those researchers
who wish to get acquainted with this fascinating topic.

\section*{Acknowledgments}

We thank Pascal Poupart for convincing us about the advantages of
SPNs, and Concha Bielza, Adnan Darwiche, Pedro Larrañaga, Alejandro
Molina, Andrzej Pronobis, Martin Trapp, Jos van de Wolfshaar, Antonio
Vergari and Kaiyu Zheng for many useful comments. However, all the
possible mistakes and omissions are our sole responsibility.

This research has been supported by grant TIN2016-77206-R from the
Spanish Government, co-financed by the European Regional Development
Fund. IP received a predoctoral grant and RSC a postdoctoral grant
from UNED, both co-financed by the Regional Government of Madrid with
funds from the Youth Employment Initiative (YEI) of the European Union.

\section*{Appendices}

\subsection*{Appendix~A\@. From arithmetic circuits to SPNs\label{sec:arithmetic-circuits}}

SPNs originated as a modification of arithmetic circuits (ACs) \cite{darwiche2003}.
Both models consist of rooted ADG with three types of nodes: sum,
product, and leaves, but a leaf node in an AC can only be a non-negative
parameter or an indicator for a Boolean variable. In contrast, the
numeric parameters of SPNs are placed in the links outgoing from sum
nodes and leaf nodes can be any univariate distribution, not only
indicators.\footnote{Section~\ref{subsec:Node-values} explains why it makes sense to
use indicators in the place of univariate distributions.} ACs are smooth (i.e., complete), decomposable, and deterministic
(i.e., selective). SPN are complete and decomposable; decomposability
might be replaced by a weaker condition, consistency \cite{poon2011},
but this generalization has no practical utility. So the main difference
between both models is that SPNs are not necessarily selective (i.e.,
deterministic), which makes them capable of representing some problems
with exponentially smaller networks than ACs \cite[Sec.~5]{choi2017},
but at the price of making much harder both exact MPE (cf.\ Sec.~\ref{subsec:MPE}
and \cite[Sec.~4]{choi2017}) and parameter learning (cf.\ Sec.~\ref{sec:Parameter-learning}).

ACs were initially proposed as an efficient representation of the
polynomials of Bayesian networks (BNs), useful for some inference
tasks \cite{darwiche2003}. Later Chavira and Darwiche \cite{chavira2007-spn-paper}
developed an alternative method for compiling BNss that may yield
exponentially smaller ACs in some cases, and Choi and Darwiche \cite{choi2017}
showed that ACs can represent arbitrary factors (called ``potentials''
in other publications), not only the conditional probability distributions
of BNs.

As we said in the introduction, the first papers about SPNs \cite{poon2011,gens2012}
presented them as representations of network polynomials of probability
distributions, just as in the case of ACs, while most posterior references,
beginning with \cite{gens2013}, present them as combinations of probability
distributions, which is a significant difference with respect to ACs.

An important novelty in the area of ACs was the algorithm by Lowd
and Domingos \cite{lowd2008} for learning ACs directly from data,
instead of compiling them from BNs. In contrast, SPNs were proposed
from their start \cite{poon2011} as a method for building probabilistic
models from data. Nowadays there are many algorithms for learning
ACs and other types of \emph{probabilistic circuits}. The references
can be found in the tutorial by Vergari et\ al.~that we recommended
at the end of Section~\ref{sec:Extensions}.\vspace{0.6ex}

\subsection*{Appendix~B. Interpretation of sum nodes as weighted averages of
conditional probabilities}

In this appendix we show that when a sum node represents a a variable~$V$,
this node can be interpreted as the average of the conditional probabilities
given~$V$, weighted by the probabilities $P(v)$. We study first
the case in which this node is the root.
\begin{prop}
\label{prop:sum-condic}Let~$\mathcal{S}$ be an SPN, whose scope
contains at least two variables, such that its root is a sum node
that represents a variable~$V$ having~$m$ values. For every~$j$,
let $n_{\sigma(j)}$ be the product node associated with value~$v_{j}$,
as in Definition~\ref{def:represents-V}. Then 
\begin{equation}
P(v_{j})=w_{i,\sigma(j)}\;.
\end{equation}
Additionally, we can define $\widetilde{\mathbf{X}}=\text{sc}(n_{i})\setminus\{V\}$,
so that if $w_{i,\sigma(j)}\neq0$ then for every configuration~$\widetilde{\mathbf{x}}$
we have
\begin{equation}
P(\widetilde{\mathbf{x}}\,|\,v_{j})=\prod_{k\in ch(\sigma(j))\setminus n_{k}\neq\mathbb{I}_{v_{j}}}S_{k}(\mathbf{x})\;.
\end{equation}
\end{prop}
This proposition is especially interesting when each child of the
root, $n_{\sigma(j)}$, has only two children, namely $\mathbb{I}_{v_{j}}$
and another node, say $n_{k(j)}$---as in Figures~\ref{fig:BN vs SPN}
and~\ref{fig:Inference SPN}. In this case the latter equation reduces
to
\begin{equation}
P(\widetilde{\mathbf{x}}\,|\,v_{j})=S_{k(j)}(\mathbf{x})\;.
\end{equation}
Given that $\text{sc}(n_{k(j)})=\widetilde{\mathbf{X}}$ for every~$j$,
Theorem~\ref{thm:Pi-Si} implies that $S_{k(j)}(\widetilde{\mathbf{x}})=P_{k(j)}(\widetilde{\mathbf{x}})$.
Therefore, the probability computed by the root,
\begin{equation}
P(\widetilde{\mathbf{x}})=S(\widetilde{\mathbf{x}})=\sum_{j}w_{i,\sigma(j)}\cdot S_{k(j)}(\widetilde{\mathbf{x}})\;,
\end{equation}
can be interpreted as a weighted average of the probabilities $P_{k(j)}(\widetilde{\mathbf{x}})$,
\begin{equation}
P(\widetilde{\mathbf{x}})=\sum_{j}w_{i,\sigma(j)}\cdot P_{k(j)}(\widetilde{\mathbf{x}})\;,
\end{equation}
 or as a particular case of the law of total probability:
\begin{equation}
P(\widetilde{\mathbf{x}})=P(\widetilde{\mathbf{x}}\,|\,v_{j})\cdot P(v_{j})\;.\label{eq:sum-condic}
\end{equation}

\begin{example}
The root node in Figure~\ref{fig:Inference SPN} represents variable~$A$,
so its children represent the probabilities $P(a)$:
\begin{align*}
P(a_{1}) & =P(+a)=w_{1,\sigma(1)}=w_{1,2}=0.3\\
P(a_{2}) & =P(\neg a)=w_{1,\sigma(2)}=w_{1,3}=0.7\;.
\end{align*}
For every $\widetilde{\mathbf{x}}\in\text{conf}^{*}(\widetilde{\mathbf{V}})=\{B,C\}$---for
example, $(+b,+c$), $(+b)$, or $(+c)$---we have
\begin{align*}
P(\widetilde{\mathbf{x}}\,|\,\text{+}a) & =\prod_{k\in ch(2)\setminus n_{k}\neq\mathbb{I}_{+a}}S_{k}(\mathbf{x})=S_{6}(\widetilde{\mathbf{x}})\\
P(\widetilde{\mathbf{x}}\,|\,\neg a) & =\prod_{k\in ch(3)\setminus n_{k}\neq\mathbb{I}_{\neg a}}S_{k}(\mathbf{x})=S_{7}(\widetilde{\mathbf{x}})
\end{align*}
and
\begin{align*}
P(\widetilde{\mathbf{x}}) & =\underset{S_{6}(\widetilde{\mathbf{x}})}{\underbrace{P(\widetilde{\mathbf{x}}\,|\,\text{+}a)}}\cdot\underset{w_{1,2}}{\underbrace{P(+a)}}+\underset{S_{7}(\widetilde{\mathbf{x}})}{\underbrace{P(\widetilde{\mathbf{x}}\,|\,\text{\textlnot}a)}}\cdot\underset{w_{1,3}}{\underbrace{P(\text{\textlnot}a)}}\;.
\end{align*}
\end{example}
If every ancestor of a sum node~$n_{i}$ represents a variable, then
the above interpretation is still valid for the context defined by
the ancestors of~$n_{i}$, $\mathbf{A}$. In this case Equation~\ref{eq:sum-condic-2}
becomes
\begin{equation}
P(\widetilde{\mathbf{x}}\,|\,\mathbf{a})=\sum_{j=1}^{m}\underset{S_{k(j)}(\widetilde{\mathbf{x}})}{\underbrace{P(\widetilde{\mathbf{x}}\,|\,v_{j},\mathbf{a})}}\cdot\underset{w_{i,\sigma(j)}}{\underbrace{P(v_{j}\,|\,\mathbf{a})}}\;.\label{eq:sum-condic-2}
\end{equation}

\begin{example}
Node~$n_{6}$ in Figure~\ref{fig:BN vs SPN} represents variable~$B$.
Its only ancestor sum node, $n_{1}$, represents variable~$A$. The
path from the root to~$n_{6}$ defines the scenario $\{A=+a\}$.
We have $\text{sc}(n_{6})=\{B,C\}$ and $\widetilde{\mathbf{V}}=\text{sc}(n_{6})\setminus\{B\}=\{C\}$.
In this example Equation~\ref{eq:sum-condic-2} instantiates into
\begin{align*}
P(c\,|\,\text{+}a)=\; & \underset{S_{14}(\widetilde{\mathbf{x}})}{\underbrace{P(c\,|\,\text{+}b,\text{+}a)}}\cdot\underset{w_{6,8}}{\underbrace{P(+b\,|\,\text{+}a)}}\:+\\
 & \underset{S_{15}(\widetilde{\mathbf{x}})}{\underbrace{P(c\,|\,\text{\textlnot}b,+a)}}\cdot\underset{w_{6,9}}{\underbrace{P(\text{\textlnot}b\,|\,\text{+}a)}}\;.
\end{align*}
Similarly, $n_{7}$ corresponds to the scenario $\{A=\neg a\}$, so
\begin{align*}
P(c\,|\,\text{\textlnot}a)=\; & \underset{S_{16}(\widetilde{\mathbf{x}})}{\underbrace{P(c\,|\,\text{+}b,\text{\textlnot}a)}}\cdot\underset{w_{7,10}}{\underbrace{P(+b\,|\,\neg a)}}\;+\\
 & \underset{S_{17}(\widetilde{\mathbf{x}})}{\underbrace{P(c\,|\,\text{\textlnot}b,\neg a)}}\cdot\underset{w_{7,11}}{\underbrace{P(\text{\textlnot}b\,|\,\neg a)}}\;.
\end{align*}
\vspace{0.3ex}
\end{example}

\subsection*{Appendix~C\@. Proofs\label{sec:Proofs}}

This appendix contains the proofs of all the propositions and the
theorem.

\begin{IEEEproof}[Proof of Proposition~\ref{prop:marginal-distr}]
It follows from Equation~\ref{eq:P-marginal} that $P(\mathbf{x})\geq0$
and
\[
\sum_{\mathbf{x}}P(\mathbf{x})=\sum_{\mathbf{x}}\sum_{\mathbf{v}\mid\mathbf{v}^{\downarrow\mathbf{X}}=\mathbf{x}}P(\mathbf{v})\;.
\]
Given that each configuration of $\mathbf{V}$ is compatible with
exactly one configuration of $\mathbf{X}$, we have
\[
\sum_{\mathbf{x}}P(\mathbf{x})=\sum_{\mathbf{v}}P(\mathbf{v})=1\;.
\]
\end{IEEEproof}
Before proving Propositions~\ref{prop:sum-Pj} and \ref{prop:prod-Pj},
we introduce a new result.
\begin{prop}
\label{prop:P-empty-conf}Let $P$ be a function $P:\text{conf}^{*}(\mathbf{V})\mapsto\mathbb{R}$
that satisfies Equation~\ref{eq:P-marginal}. If $P(\blacklozenge)=1$,
then $\sum_{\mathbf{v}}P(\mathbf{v})=1$.
\end{prop}
\begin{IEEEproof}
Because of Equation~\ref{eq:P-marginal}, with $\mathbf{X}=\varnothing$,
we have
\[
P(\blacklozenge)=\sum_{\mathbf{v}\mid\mathbf{v}^{\downarrow\varnothing}=\blacklozenge}P(\mathbf{v})\;.
\]
Taking into account that the empty configuration is compatible with
every configuration of every set,
\[
P(\blacklozenge)=\sum_{\mathbf{v}}P(\mathbf{v})\;.
\]
\end{IEEEproof}
\begin{IEEEproof}[Proof of Proposition~\ref{prop:sum-Pj}]
We first prove that~$P$ satisfies Equation~\ref{eq:P-marginal}
and then that it is a probability distribution:
\begin{align*}
P(\mathbf{x}) & =\sum_{j=1}^{n}w_{j}\cdot P_{j}(\mathbf{x})=\sum_{j=1}^{n}w_{j}\cdot\sum_{\mathbf{v}\mid\mathbf{v}^{\downarrow\mathbf{X}}=\mathbf{x}}P_{j}(\mathbf{v})\\
 & =\sum_{\mathbf{v}\mid\mathbf{v}^{\downarrow\mathbf{X}}=\mathbf{x}}\sum_{j=1}^{n}w_{j}\cdot P_{j}(\mathbf{v})=\sum_{\mathbf{v}\mid\mathbf{v}^{\downarrow\mathbf{X}}=\mathbf{x}}P(\mathbf{v})\;,
\end{align*}
It is clear that $P(\mathbf{v})\geq0$ for all~$\mathbf{v}\in$ conf$(\mathbf{V})$
and, because of Proposition~\ref{prop:P-empty-conf},
\[
\sum_{\mathbf{v}}P(\mathbf{v})=P(\mathbf{\blacklozenge})=\sum_{j=1}^{n}w_{j}\cdot P_{j}(\mathbf{\mathbf{\blacklozenge}})=\sum_{j=1}^{n}w_{j}=1\;.
\]
\end{IEEEproof}
\begin{IEEEproof}[Proof of Proposition~\ref{prop:prod-Pj}]
When $n=1$ the proof is trivial because $P=P_{1}$. When $n=2$
we have, for every configuration~$\mathbf{x}$, with $\mathbf{X}\subseteq\mathbf{V}$,
\begin{align*}
\sum_{\mathbf{v}\mid\mathbf{v}^{\downarrow\mathbf{X}}=\mathbf{x}}P(\mathbf{v}) & =\sum_{\mathbf{v}\mid\mathbf{v}^{\downarrow\mathbf{X}}=\mathbf{x}}P_{1}(\mathbf{v}^{\downarrow\mathbf{V}_{1}})\cdot P_{2}(\mathbf{v}^{\downarrow\mathbf{V}_{2}})\;.
\end{align*}
Given that $\mathbf{V}_{1}\cup\mathbf{V}_{2}=\mathbf{V}$ and $\mathbf{V}_{1}\cap\mathbf{V}_{2}=\varnothing$,
every configuration of~$\mathbf{V}$ can be obtained from a composition
of two configurations, $\mathbf{v}=\mathbf{v}_{1}\mathbf{v}_{2}$,
where $\mathbf{v}_{1}=\mathbf{v}^{\downarrow\mathbf{V}_{1}}$ and
$\mathbf{v}_{2}=\mathbf{v}^{\downarrow\mathbf{V}_{1}}$. The condition
$\mathbf{v}^{\downarrow\mathbf{X}}=\mathbf{x}$---the compatibility
of $\mathbf{x}$ with~$\mathbf{v}$---can be decomposed into two
conditions: $\mathbf{v}_{1}^{\downarrow\mathbf{X}_{1}}=\mathbf{x}^{\downarrow\mathbf{X}_{1}}$
and $\mathbf{v}_{2}^{\downarrow\mathbf{X}_{2}}=\mathbf{x}^{\downarrow\mathbf{X}_{2}}$,
where $\mathbf{X}_{1}=\mathbf{V}_{1}\cap\mathbf{X}$ and $\mathbf{X}_{2}=\mathbf{V}_{2}\cap\mathbf{X}$
. Therefore, 
\[
\sum_{\mathbf{v}\mid\mathbf{v}^{\downarrow\mathbf{X}}=\mathbf{x}}P(\mathbf{v})=\sum_{\mathbf{v}_{1}\mid\mathbf{v}^{\downarrow\mathbf{X}_{1}}=\mathbf{x}^{\downarrow\mathbf{X}_{1}}}\sum_{\mathbf{v}_{2}\mid\mathbf{v}^{\downarrow\mathbf{X}_{2}}=\mathbf{x}^{\downarrow\mathbf{X}_{2}}}P_{1}(\mathbf{v}_{1})\cdot P_{2}(\mathbf{v}_{2})\;.
\]
The property $\mathbf{V}_{1}\cap\mathbf{V}_{2}=\varnothing$ also
implies that $P_{1}(\mathbf{v}_{1})$ does not depend on~$\mathbf{v}_{2}$
and vice versa, so
\begin{align*}
\sum_{\mathbf{v}\mid\mathbf{v}^{\downarrow\mathbf{X}}=\mathbf{x}}P(\mathbf{v}) & =\left(\sum_{\mathbf{v}_{1}\mid\mathbf{v}_{1}=\mathbf{x}^{\downarrow\mathbf{X}_{1}}}P_{1}(\mathbf{v}_{1})\right)\cdot\sum_{\mathbf{v}_{2}\mid\mathbf{v}_{2}=\mathbf{x}^{\downarrow\mathbf{X}_{2}}}P_{2}(\mathbf{v}_{2})\;.
\end{align*}
The fact that $P_{1}$ is a probability function implies that
\[
P_{1}(\mathbf{x}^{\downarrow\mathbf{X}_{1}})=\sum_{\mathbf{v}_{1}\mid\mathbf{v}_{1}=\mathbf{x}^{\downarrow\mathbf{X}_{1}}}P_{1}(\mathbf{v}_{1})
\]
and the definition $\mathbf{X}_{1}=\mathbf{V}_{1}\cap\mathbf{X}$
implies that $\mathbf{x}^{\downarrow\mathbf{X}_{1}}=\mathbf{x}^{\downarrow\mathbf{V}_{1}}$.
Therefore
\[
\sum_{\mathbf{v}\mid\mathbf{v}^{\downarrow\mathbf{X}}=\mathbf{x}}P(\mathbf{v})=P_{1}(\mathbf{x}^{\downarrow\mathbf{V}_{1}})\cdot P_{2}(\mathbf{x}^{\downarrow\mathbf{V}_{2}})=P(\mathbf{x})\;,
\]
which proves that~$P$ satisfies Equation~\ref{eq:P-marginal}.
It is clear that $P(\mathbf{v})\geq0$ for all~$\mathbf{v}\in$ conf$(\mathbf{V})$
and, because of Proposition~\ref{prop:P-empty-conf},
\[
\sum_{\mathbf{v}}P(\mathbf{v})=P(\mathbf{\blacklozenge})=P_{1}(\mathbf{\blacklozenge})\cdot P_{2}(\mathbf{\blacklozenge})=1\;,
\]
which completes the proof for $n=2$. As a consequence, if $P_{1}(\mathbf{x}^{\downarrow\mathbf{V}_{1}})\cdot\ldots\cdot P_{n-1}(\mathbf{x}^{\downarrow\mathbf{V}_{n-1}})$
and $P_{n}(\mathbf{x}^{\downarrow\mathbf{V}_{n}})$ are probability
functions for disjoint $\mathbf{V}_{j}$'s, then $P_{1}(\mathbf{x}^{\downarrow\mathbf{V}_{1}})\cdot\ldots\cdot P_{n}(\mathbf{x}^{\downarrow\mathbf{V}_{n}})$
is also a probability function, which proves Proposition~\ref{prop:prod-Pj}
for any value of $n$.
\end{IEEEproof}

\begin{IEEEproof}[Proof of Proposition~\ref{prop:decomp-node}]
Let~$n_{j}$ and~$n_{j'}$ be two different children of~$n_{i}$.
If $\textit{descendants}(n_{j})\cap\textit{descendants}(n_{j'})=\varnothing$,
then $\textit{sc}(n_{j})\cap\textit{sc}(n_{j'})=\varnothing$, i.e.,
their scopes are disjoint.

Reciprocally, if a node~$n_{k}$ is a descendant of both~$n_{j}$
and~$n_{j'}$ then~$\textit{sc}(n_{k})\subseteq\textit{sc}(n_{j})\cap\textit{sc}(n_{j'})$,
which implies that the scopes are not disjoint because, by the definition
of scope, $\textit{sc}(n_{k})\neq\varnothing$.
\end{IEEEproof}
\begin{IEEEproof}[Proof of Theorem~\ref{thm:Pi-Si}]
Let~$n_{i}$ be a terminal node representing $P(v)$. In this case
the set~$\mathbf{V}$ in Definition~\ref{def:prob-function} is
$\{V\}$, and $P_{i}(v)=S_{i}(v)=P(v)$ is a probability distribution.
If $\mathbf{X}\subset\mathbf{V}$ then $\mathbf{X}=\varnothing$ and
Equation~\ref{eq:P-marginal} holds because, on the one hand, $P_{i}(\blacklozenge)=S_{i}(\blacklozenge)=1$
(Eqs.~\ref{eq:Si-leaf} and~\ref{eq:Si=00003DPi}) and, on the other,
\[
\sum_{\mathbf{v}\mid\mathbf{v}^{\downarrow\varnothing}=\mathbf{\blacklozenge}}P_{i}(\mathbf{v})=\sum_{v}P_{i}(v)=1\ .
\]

Let $n_{i}$ be a non-terminal node. We assume that~$P_{j}$ is a
probability distribution for each of its children, $n_{j}$. If~$n_{i}$
is a sum node then $P_{i}$ is a probability functiojn because of
Proposition~\ref{prop:sum-Pj}, with $\mathbf{V}=\text{sc}(n_{i})=\text{sc}(n_{j})$.
The completeness of the SPN guarantees that $n_{i}$ and all its children
have the same scope. If~$n_{i}$ is a product node then $P_{i}$
is a probability function because of Proposition~\ref{prop:prod-Pj},
with $\mathbf{V}_{j}=\text{sc}(n_{j})$. The decomposability of the
SPN guarantees that the $\mathbf{V}_{j}$'s are disjoint and the definition
of sc$(n_{i})$ ensures that $\bigcup_{j}\mathbf{V}_{j}=\mathbf{V}=\text{sc}(n_{i})$.
\end{IEEEproof}
The following example shows that if an SPN is not complete it may
overestimate the probability of some configurations.
\begin{example}
Let~${\cal S}$ be an SPN whose root, $n_{0}$, is a sum node with
two children, $n_{1}$ and~$n_{2}$, such that $w_{0,1}=0.6$ and
$w_{0,2}=0.4$. These children represent two probability distributions,
$P_{1}$ and~$P_{2}$, defined on~$V_{1}$ and~$V_{2}$, respectively,
with $P_{1}(+v_{1})=P_{1}(\neg v_{1})=P_{2}(+v_{2})=P_{2}(\neg v_{2})=0.5.$
Then $S(+v_{1})=w_{0,1}\cdot S_{1}(+v_{1})+w_{0,2}\cdot S_{2}(+v_{1})=0.6\cdot0.5+0.4\cdot1=0.7$
and $S(\neg v_{1})=0.7$. $S(\mathbf{x})$ is not a probability function
because $S(+v_{1})+S(\neg v_{1})>1$.
\end{example}
The following example shows that if an SPN is not decomposable it
may underestimate the probability of some configurations.
\begin{example}
Let~${\cal S}$ be an SPN whose root, $n_{0}$, is a product node
with two children, $n_{1}$ and~$n_{2}$, which represent two probability
distributions, $P_{1}$ and~$P_{2}$ respectively, both defined on~$V$,
with $P_{1}(+v)=P_{1}(\neg v)=P_{2}(+v)=P_{2}(\neg v)=0.5.$ Then
$S(+v)=S_{1}(+v)\cdot S_{2}(+v)=0.5\cdot0.5=0.25$ and $S(\neg v)=0.25$.
$S(\mathbf{x})$ is not a probability function because $S(+v)+S(\neg v)<1$.
In this case it is not possible to invoke Proposition~\ref{prop:prod-Pj}
because~$P_{1}$ and~$P_{2}$ are not defined on disjoint subsets
of variables.

This is the reason why SPNs are required to be complete and decomposable.
\end{example}
\begin{IEEEproof}[Proof of Proposition~\ref{prop:represents-V}]
Let~$\mathbf{v}\in\text{conf}(\mathcal{S})$ and~$j^{*}$ the integer
in $\{1,\ldots,m\}$ such that $v_{j^{*}}=\mathbf{v}^{\downarrow V}$.
Let $j\in\{j\text{1,\ensuremath{\ldots,m\}}}$ with $j\neq j^{*}$;
then $\mathbb{I}_{v_{j}}(\mathbf{v})=0$ because $v_{j}\neq\mathbf{v}^{\downarrow V}$.
If $n_{\sigma(j)}=\mathbb{I}_{v_{j}}$ (the indicator is a child of~$n_{i}$)
then $S_{\sigma(j)}(\mathbf{v})=\mathbb{I}_{v_{j}}(\mathbf{v})=0$.
If $n_{\sigma(j)}=\mathbb{I}_{v_{j}}$ (the indicator is a grandchild
of~$n_{i}$) then~$n_{\sigma(j)}$ is a product node and the contributions
of its other children are multiplied by~0, which implies that $S_{\sigma(j)}(\mathbf{v})=0$.
\end{IEEEproof}
\begin{IEEEproof}[Proof of Proposition~\ref{prop:augmented-SPN}]
As the augmentation processes the nodes one by one, it suffices to
prove the proposition for the SPN that results from augmenting one
non-selective node, $n_{i}$. The addition of the indicators for~$Z$
makes this node selective in~${\cal S}'$ because of Definition~\ref{def:represents-V}
and Proposition~\ref{prop:represents-V}, without affecting its completeness.
The addition of~$n_{k}'$ makes all its ancestors complete and decomposable,
without affecting selectivity---see Figure~\ref{fig:augmented-SPN}.

Let $\mathbf{x}\in\text{conf}^{*}(\mathcal{S})$. Given that $Z\notin\text{\emph{sc}}(\mathcal{S}$),
$\mathbb{I}_{z(j)}=1$ for every~$j$. If we denote by $S_{h}'(\mathbf{x})$
the value of node~$n_{h}$ in~${\cal S}'$, we have $S_{j}'(\mathbf{x})=S_{j}(\mathbf{x})$,
$S_{j'}'(\mathbf{x})=S_{j'}(\mathbf{x})$, $S_{i}'(\mathbf{x})=S_{i}(\mathbf{x})$,
$S_{k}'(\mathbf{x})=S_{k}(\mathbf{x})$, $S_{i'}'(\mathbf{x})=1$,
$S_{k'}'(\mathbf{x})=S_{k}(\mathbf{x})$, $S_{l}'(\mathbf{x})=S_{l}(\mathbf{x})$,
etc. Therefore, for any node~$n_{h}\in{\cal S}$ we have $\,S_{h}'(\mathbf{x})=S_{h}(\mathbf{x})$
and, consequently, $P'(\mathbf{x})=P(\mathbf{x})$.
\end{IEEEproof}
\begin{IEEEproof}[Proof of Proposition~\ref{prop:S_v(v)}]
${\cal S}_{\mathbf{v}}$ contains all the paths making positive contributions
to $S(\mathbf{v})$.
\end{IEEEproof}
\begin{IEEEproof}[Proof of Proposition~\ref{prop:induced-tree}]
Selectivity implies that if~$n_{i}$ is a sum node in~$\mathcal{S}$
and $S_{i}(\mathbf{v})\neq0$, then it has exactly one child~$n_{j}$
such that $S_{j}(\mathbf{v})\neq0$ and $w_{ij}>0$ (which implies
that this link has not been removed), so in~$\mathcal{S}_{\mathbf{v}}$
every sum node~$n_{i}$ has exactly one child. Every node in~${\cal S}_{\mathbf{v}}$
other than the root has at least one parent---otherwise, it would
have been removed. We now prove that no node can have more than one
parents. Let us assume that a certain node~$n_{j}$ has two parents,~$n_{i}$
and~$n_{i'}$. They have at least one common ancestor, the root.
Let~$n_{k}$ an ancestor of both~$n_{i}$ and~$n_{i'}$ such that
no descendant of~$n_{k}$ is an ancestor of both of them. Node~$n_{k}$
cannot be a sum node because each sum node has only one child. It
cannot be a product node either because then~$n_{k}$ would have
two children with a common descendant, $n_{j}$, which is impossible
in a decomposable SPN (cf.\ Prop.~\ref{prop:decomp-node}). Therefore
assuming that a node has two or more parents leads to a contradiction.
\end{IEEEproof}
\begin{IEEEproof}[Proof of Proposition~\ref{prop:prod-w_ij}]
The proposition holds trivially when the SPN contains only one node.
Let us consider a non-terminal node~$n_{i}$. For each child~$n_{j}$
the sub-SPN rooted at~$n_{j}$ is also selective. We assume that
the proposition holds for each of these sub-SPNs. If~$n_{i}$ is
a sum node, the proposition holds for~$n_{i}$ because it only has
only one child in ${\cal T}_{\mathbf{v}}$ and $S_{\mathbf{v}}(\mathbf{v})=S(\mathbf{v})$.
If~$n_{i}$ is a product node, the proposition follows from Equation~\ref{eq:Si-prod}.
\end{IEEEproof}
\begin{IEEEproof}[Proof of Corollary~\ref{cor:prod-w_ij}]
${\cal T}_{\mathbf{v}}$ only contains the indicators for which $S_{k}(\mathbf{v})=1$.
\end{IEEEproof}
\begin{IEEEproof}[Proof of Proposition~\ref{prop:nij-log-wij}]
Equation~\ref{eq:prod-w_ij} implies that
\begin{equation}
\log S(\mathbf{v}^{t}\,|\,\mathbf{w})=\sum_{(i,j)\in\mathcal{T}_{\mathbf{v}^{t}}}\log w_{ij}+c_{t}\;,\label{eq:log-S(vt|w)}
\end{equation}
where
\[
c_{t}=\begin{cases}
\log S_{k}(\mathbf{v}^{t}) & \text{if }n_{k}\text{ is terminal in }{\cal T}_{\mathbf{v}^{t}}\\
0 & \text{otherwise}\;.
\end{cases}
\]
Whether~$n_{k}$ is terminal in~${\cal T}_{\mathbf{v}^{t}}$ does
not depend on the actual values of the weights, provided that they
are all different from~0. If we define
\[
c=\sum_{t=1}^{T}c_{t}
\]
we have 
\begin{align*}
L_{\mathcal{D}}(\mathbf{w}) & =\sum_{t=1}^{T}\log S(\mathbf{v}^{t}\,|\,\mathbf{w})=\sum_{w_{ij}\in\mathbf{W}}n_{ij}\cdot\log w_{ij}+c\;,
\end{align*}
where $n_{ij}$ is the number of instances in the dataset for which
$(i,j)\in\mathcal{T}_{\mathbf{v}^{t}}$ and~$c$ does not depend
on~$\mathbf{w}$.
\end{IEEEproof}
Now, before proving Proposition~\ref{prop:nij-EM}, we introduce
an auxiliary result, adapted from \cite{darwiche2003}.
\begin{prop}
\label{prop:der-Sv-wij}If $\mathcal{S}$ is selective, $\mathbf{v}\in\text{conf}(\mathbf{\mathcal{S}})$,
$S(\mathbf{v})\neq0$, and $(i,j)\in\mathcal{S_{\mathbf{v}}}$, then
\begin{equation}
S(\mathbf{v})=w_{ij}\cdot\frac{\partial S(\mathbf{v})}{\partial w_{ij}}\;.
\end{equation}
\end{prop}
\begin{IEEEproof}
Equation~\ref{eq:prod-w_ij} implies $w_{ij}\neq0$ (because $S(\mathbf{v})\neq0$)
and
\[
\frac{\partial S(\mathbf{v})}{\partial w_{ij}}=\frac{S(\mathbf{v})}{w_{ij}}\;.
\]
\end{IEEEproof}
\begin{IEEEproof}[Proof of Proposition~\ref{prop:nij-EM}]
Given that $\mathbf{v}^{t}\mathbf{h}^{t}$ is a complete configuration
of $\text{sc}({\cal S}')$, i.e., $\mathbf{v}^{t}\mathbf{h}^{t}\in\text{conf}({\cal S}')$,
and~${\cal S}'$ is selective, Proposition~\ref{prop:der-Sv-wij}
implies that
\[
S'(\mathbf{v}^{t}\mathbf{h}^{t})=w_{ij}\cdot\frac{\partial S'(\mathbf{v}^{t}\mathbf{h}^{t})}{\partial w_{ij}}
\]
and
\[
P'(\mathbf{h}^{t}\,|\,\mathbf{v}^{t})=\frac{S'(\mathbf{v}^{t}\mathbf{h}^{t})}{S'(\mathbf{v}^{t})}=w_{ij}\cdot\frac{1}{S'(\mathbf{v}^{t})}\cdot\frac{\partial S'(\mathbf{v}^{t}\mathbf{h}^{t})}{\partial w_{ij}}\;.
\]
So Equation~\ref{eq:nij-EM-1} can be rewritten as 
\[
n_{ij}=\sum_{t=1}^{T}w_{ij}\cdot\frac{1}{S'(\mathbf{v}^{t})}\cdot\sum_{\mathbf{h}^{t}\,|\,(i,j)\in{\cal T}'_{\mathbf{v}^{t}\mathbf{h}^{t}}}\frac{\partial S'(\mathbf{v}^{t}\mathbf{h}^{t})}{\partial w_{ij}}\;.
\]
If link $(i,j)$ does not belong to the tree induced by $\mathbf{v}^{t}\mathbf{h}^{t}$
then $\partial S'(\mathbf{v}^{t}\mathbf{h}^{t})/\partial w_{ij}=0$,
so the inner summation in the previous expression can be extended
to all the configurations of~$\mathbf{H}^{t}$:
\[
n_{ij}=\sum_{t=1}^{T}w_{ij}\cdot\frac{1}{S'(\mathbf{v}^{t})}\cdot\sum_{\mathbf{h}^{t}}\frac{\partial S'(\mathbf{v}^{t}\mathbf{h}^{t})}{\partial w_{ij}}\;.
\]
Given that $S'(\mathbf{v}^{t})=\sum_{\mathbf{h}^{t}}S'(\mathbf{v}^{t}\mathbf{h}^{t})$,
we have
\[
n_{ij}=\sum_{t=1}^{T}w_{ij}\cdot\frac{1}{S'(\mathbf{v}^{t})}\cdot\frac{\partial S'(\mathbf{v}^{t})}{\partial w_{ij}}\;.
\]
Because of Proposition~\ref{prop:augmented-SPN}, $S'(\mathbf{v}^{t})=S(\mathbf{v}^{t})$
and
\begin{align*}
n_{ij} & =\sum_{t=1}^{T}w_{ij}\cdot\frac{1}{S(\mathbf{v}^{t})}\cdot\frac{\partial S(\mathbf{v}^{t})}{\partial w_{ij}}\\
 & =\sum_{t=1}^{T}w_{ij}\cdot\frac{1}{S(\mathbf{v}^{t})}\cdot\frac{\partial S(\mathbf{v}^{t})}{\partial S_{i}}\cdot\frac{\partial S_{i}(\mathbf{v}^{t})}{\partial w_{ij}}\;.
\end{align*}
This result, together with Equations~\ref{eq:S-deriv-i} and~\ref{eq:Si-sum},
leads to Equation~\ref{eq:nij-EM-2}.
\end{IEEEproof}
\begin{IEEEproof}[Proof of Proposition~\ref{prop:sum-condic}]
Let $j\in\{1,\ldots m\}$. Since~$n_{r}$ represents variable~$V$,
either $\mathbb{I}_{v_{j}}=n_{\sigma(j)}$ or~$\mathbb{I}_{v_{j}}$
is a child of~$n_{\sigma(j)}$---see Definition~\ref{def:represents-V}.
In the first case, we would have $\text{sc}(n_{\sigma(j)})=\{V\}$
and the completeness of~$\mathcal{S}$ would imply that $\text{sc}(n_{r})=\text{sc}(n_{r})=\{V\}$,
in contradiction with the assumption that~$\text{sc}(\mathcal{S})$
has at least to variables. Therefore~$\mathbb{I}_{v_{j}}$ must be
a child of~$n_{\sigma(j)}$, a product node, and
\[
S_{\sigma(j)}(\mathbf{x})=\mathbb{I}_{v_{j}}(\mathbf{x})\cdot\prod_{k\in ch(\sigma(j))\setminus n_{k}\neq\mathbb{I}_{v_{j}}}S_{k}(\mathbf{x})\;,
\]
\[
\]
for every $\mathbf{x\in\text{conf}^{*}(\mathcal{S})}$. Let $\mathbf{x}=v_{j}\mathbf{\widetilde{\mathbf{x}}}$,
i.e., the composition of~$v_{j}$ and any $\widetilde{\mathbf{x}}\in\text{conf}^{*}(\widetilde{\mathbf{V}})$.
When $j'\neq j$ we have $\mathbb{I}_{v_{j}}(v_{j}\mathbf{\widetilde{\mathbf{x}}})=\mathbb{I}_{v_{j}'}(v_{j})=0$
and $S_{\sigma(j')}(v_{j}\mathbf{\widetilde{\mathbf{x}}})=0$. On
the other hand, 
\begin{align*}
S_{\sigma(j)}(v_{j}\mathbf{\widetilde{\mathbf{x}}}) & =\mathbb{I}_{v_{j}}(v_{j})\cdot\prod_{k\in ch(\sigma(j))\setminus n_{k}\neq\mathbb{I}_{v_{j}}}S_{k}(\mathbf{\widetilde{\mathbf{x}}})\;,\\
 & =\prod_{k\in ch(\sigma(j))\setminus n_{k}\neq\mathbb{I}_{v_{j}}}S_{k}(\mathbf{\widetilde{\mathbf{x}}})\;.
\end{align*}
Since $n_{r}$ is a sum node, 
\begin{align*}
P(v_{j}\mathbf{\widetilde{\mathbf{x}}}) & =\sum_{j'=1}^{m}w_{i,\sigma(j')}\cdot S_{\sigma(j')}(v_{j}\mathbf{\widetilde{\mathbf{x}}})\\
 & =w_{i,\sigma(j)}\cdot\prod_{k\in ch(\sigma(j))\setminus n_{k}\neq\mathbb{I}_{v_{j}}}S_{k}(\mathbf{\widetilde{\mathbf{x}}})\;.
\end{align*}
In particular, if $\widetilde{\mathbf{x}}=\blacklozenge$ then $S_{k}(\blacklozenge)=1$
for every~$k$ and
\[
P(v_{j})=w_{i,\sigma(j)}\;.
\]
If $w_{i,\sigma(j)}\neq0$,
\[
P(v_{j}\,|\,\mathbf{\widetilde{\mathbf{x}}})=\frac{P(v_{j}\mathbf{\widetilde{\mathbf{x}}})}{P(v_{j})}=\prod_{k\in ch(\sigma(j))\setminus n_{k}\neq\mathbb{I}_{v_{j}}}S_{k}(\mathbf{\widetilde{\mathbf{x}}})\;.
\]
\end{IEEEproof}
\bibliographystyle{ieeetr}
\bibliography{SPN,cisiad}

\end{document}